\documentclass[journal]{IEEEtran}

\usepackage{verbatim}
\usepackage{amsmath}

\usepackage[utf8]{inputenc}
\usepackage{graphicx}
\usepackage{subfigure}
\usepackage{float}
\usepackage{bm}
\usepackage{amssymb}
\usepackage{threeparttable}
\usepackage{multirow}
\usepackage{booktabs}

\usepackage{graphics} 
\usepackage{epsfig} 
\usepackage[T1]{fontenc}
\usepackage{cite}
\usepackage{tabularx}
\usepackage{color}
\usepackage[colorlinks=true]{hyperref}

\graphicspath{{Images/}}

\title{Learning to Assist Different Wearers in Multitasks: 
Efficient and Individualized Human-In-the-Loop Adaption Framework for Exoskeleton Robots 
}
\author{Yu Chen, Gong Chen, Jing Ye, Chenglong Fu, Bin Liang, and Xiang Li
\thanks{Y. Chen, B. Liang, and X. Li are with the Department of Automation, Tsinghua University, China. G. Chen and J. Ye are with the Shenzhen MileBot Robotics Co., Ltd, China. C. Fu is with the Department of Mechanical and Energy Engineering, Southern University of Science and Technology, Shenzhen 518055, China. This work was supported in part by the Science and Technology Innovation 2030-Key Project under Grant 2021ZD0201404, in part by
the Institute for Guo Qiang, Tsinghua University, and in part by the National Natural Science Foundation of China under Grant U21A20517 and 52075290. 
Corresponding author: Xiang Li (xiangli@tsinghua.edu.cn)}%
}

\begin{document}

\maketitle
\pagestyle{empty}  
\thispagestyle{empty} 

\begin{abstract}
One of the typical purposes of using lower-limb exoskeleton robots is to provide assistance to the wearer by supporting their weight and augmenting their physical capabilities according to a given task and human motion intentions. 
The generalizability of robots across different wearers in multiple tasks is important to ensure that the robot can provide correct and effective assistance in actual implementation.
However, most lower-limb exoskeleton robots exhibit only limited generalizability.
Therefore, this paper proposes a human-in-the-loop learning and adaptation framework for exoskeleton robots to improve their performance in various tasks and for different wearers. 
To suit different wearers, an individualized walking trajectory is generated online using dynamic movement primitives and Bayes optimization.
To accommodate various tasks, a task translator is constructed using a neural network to generalize a trajectory to more complex scenarios.
These generalization techniques are integrated into a unified variable impedance model, which regulates the exoskeleton to provide assistance while ensuring safety. In addition, an anomaly detection network is developed to quantitatively evaluate the wearer's comfort, which is considered in the trajectory learning procedure and contributes to the relaxation of conflicts in impedance control.  
The proposed framework is easy to implement, because it requires proprioceptive sensors only to perform and deploy data-efficient learning schemes.
This makes the exoskeleton practical for deployment in complex scenarios, accommodating different walking patterns, habits, tasks, and conflicts. 
Experiments and comparative studies on a lower-limb exoskeleton robot are performed to demonstrate the effectiveness of the proposed framework.
\end{abstract}

\section{Introduction}

The primary objective of a lower-limb exoskeleton robot is to support the body weight of the wearer and enhance their physical capabilities based on a given task and human motion intentions.
Such robots have been widely used in the domains of the industry \cite{de2016exoskeletons}, military \cite{zoss2006biomechanical}, and rehabilitation \cite{chen2019elbow}. 
Various prototypes and even commercialized products, e.g., ReWalk \cite{esquenazi2017powered}, Ekso \cite{brenner2016exploring} and HAL \cite{iizuka2021regulation}, have been developed to address different needs.
Moreover, a range of control approaches have been developed to allow the robot to provide empathetic assistance\cite{teramae2017emg,pehlivan2015minimal}. 
However, most of the current lower-limb exoskeletons are designed for a specific task (e.g., ground walking only) or personalized for individual users. 
When they are deployed for different tasks or users, manual programming or even fine-tuning is typically necessary to set the assistance strategy or reference trajectory.

\begin{figure}[t]
    \centering
    \includegraphics[width=8.5cm]{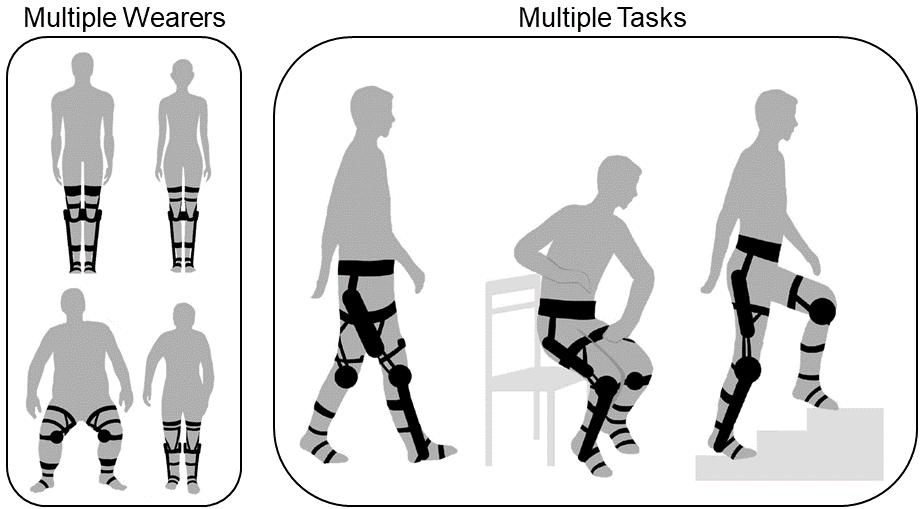}
    \caption{The proposed method enables the lower-limb exoskeleton (in black) to provide individualized assistance for multiple wearers with different backgrounds (left) and in multiple tasks (right), e.g., walking, sit-to-stand, climbing stairs.}
    \label{head}
\end{figure}

Although the tasks of lower-limb exoskeleton in rehabilitation training are well-defined and largely invariable, 
the demands are more diverse when assisting a healthy individual (Fig.\ref{head}), encompassing various tasks (e.g., walking, squatting, standing up, and climbing stairs), different wearers (with different walking speeds, patterns, or habits), and the appearance of anomalous events during the task (e.g., physical conflicts, fatigue, and imbalance). 
Therefore, the ability of online learning and adaptation to multiple tasks and wearers is important to ensure that the robot provides effective assistance in practical applications.

To fulfill these requirements, a novel human-in-the-loop (HIL) framework is established in this study to enhance the online learning and adaptation ability of lower-limb exoskeletons.
The contributions of this work can be summarized as follows.

\begin{itemize}
\item[-] An unsupervised detection network is constructed to address anomalies during the task. 
This network prevents the robot from learning or providing assistance based on incorrect patterns without affecting its generalizability, while optimizing trajectory generation to minimize conflicts.

\item[-] An online individualizing approach is developed, using dynamic movement primitives (DMP) to parameterize the assistance trajectory and Bayesian optimization to minimize the cost function, enabling the provision of personalized assistance to multiple wearers.

\item[-] A neural network (NN) is trained to serve as the task translator.
This allows optimized assistance for ground walking to be transferred to a variety of tasks may not be suitable for online learning, such as stair climbing.

\item[-] A unified variable impedance model is proposed to integrate the individualized and generalized trajectories. 
Moreover, a singular-perturbation approach is developed to implement the model with low-computation complexity without the need for force sensors.

\end{itemize}

The proposed framework relies solely on the proprioceptive sensors of exoskeleton robots (e.g., encoders and strain gauges), without external sensors (e.g., electroencephalographic/electromyographic (EMG) sensors or motion capture systems).
Thus, it can be easily implemented, consistent with the practical scenarios of human assistance.
All phases of optimization and inference are realized online, ensuring data efficiency without the need for pre-training. 
The stability of the compliantly-driven exoskeleton and convergence to the variable impedance model are theoretically grounded. 
A series of ablation studies and experiments are conducted on a lower-limb exoskeleton robot to illustrate the effectiveness of the developed framework in different tasks and for multiple wearers.
In previous work \cite{zhang2023multi}, we developed a variable impedance controller to relax physical conflicts during the human-robot interaction by using proprioceptive sensors and a variational autoencoder (VAE) based detection network.  
In this study, we extend this controller to an anomaly-sensitive online adaptation framework, considering multiple tasks and individualization.

\section{Related Works}

The section provides an overview of research related to trajectory generation and interaction control for exoskeleton robots.\\

\noindent\textbf{Trajectory Generation}:
This study focuses on the trajectory generation for exoskeletons designed to assist multiple wearers across different tasks. 

One approach to achieving individualized assistance for multiple wearers is HIL optimization, where the body features and walking habits of the users are implicitly considered \cite{slade2022personalizing}. 
Leveraging sampling-based nonlinear optimization, a HIL optimization technique for an ankle exoskeleton was first proposed in \cite{HIL_first}, where the metabolic rate was reduced through a covariance matrix adaptation evolution strategy (CMA-ES). Further individualization was achieved through an intuitive gait parameter adjustment method named "Iterative Learning of Human Behavior (ILHB)", which optimizes joint-reference trajectories based on observed ground contact timing \cite{park2022iterative}. 
The HIL optimization concept has also been applied to wearable robots with multiple joints \cite{franks2022effects}, prostheses \cite{wen2017automatically}, and soft exosuits \cite{ding2018human}. In addition, reinforcement learning (RL) \cite{zhang2019adaptive} and impedance control \cite{li2022human} have been integrated into the HIL optimization framework to offer timely and adaptive assistance. 
Unlike these studies, which emphasized the reduction of the wearers' metabolic rate, this paper proposes a new cost function to balance the assistance and wearer's comfort.
Furthermore, instead of using multiple sensors \cite{gordon2022human,tan2022time}, the proposed learning framework relies only on proprioceptive sensors (i.e., encoders on the robot), boosting its portability and practicability.

Second, 
the operational flexibility of robots can be enhanced by improving their adaptability to different tasks. 
As the assistance profiles for different tasks vary significantly, multiple control algorithms have been designed for specific tasks, such as walking \cite{park2022iterative}, ascending stairs \cite{hood2022powered}, and addressing gait asymmetry \cite{qian2022adaptive}.
In \cite{reznick2020predicting}, the joint kinematics was simultaneously generalized for different walking speeds and slopes using interpolation techniques. 
Furthermore, kinematics of ascending stairs and transitional motion was predicted based on Bernstein basis polynomials in \cite{cheng2022modeling}, promoting robustness in the lower-limb exoskeleton.
However, these studies typically focused on a single task 
or task generalization under different configurations. In contrast, this paper proposes a new task translator to unify trajectories across multiple tasks.\\

\noindent\textbf{Interaction Control}: 
The purpose of interaction control is to drive the exoskeleton robot along the generated trajectory while ensuring the safety of the wearer. Among different control schemes, variable impedance control (VIC) has been extensively implemented in various robotic systems involving physical interaction\cite{cheah1998learning}.
VIC is inspired by human behavior and biomechanics, wherein the central nervous system can continuously modulate the physical impedance of human limbs to adapt to uncertain environments, different types of tasks, and external perturbations \cite{damm2008physiological,sharifi2021impedance}. 
In \cite{huo2021intention}, VIC was deployed in a lower-limb rehabilitation exoskeleton.
The deviation between the human intended motion and current joint angle of the exoskeleton was used to vary the impedance parameters to resolve physical conflict. In \cite{rozo2016learning}, VIC was applied to a collaborative assembly task, where the robot varied its stiffness based on the interaction with the human subject and the learned impedance behavior from demonstrations.
Furthermore, the impedance parameters can be adjusted while 
considering other variables, such as the velocity \cite{ficuciello2015variable} and center of mass location \cite{pai2000thresholds}. 

Several learning-based approaches have also been developed to achieve interaction control.
In \cite{zhang2021learning}, an inverse reinforcement learning-based method was designed for impedance learning, where the reward function associated with the trajectory tracking error and impedance switching policy for the action space were learned from expert demonstrations. 
In \cite{bogdanovic2020learning}, a reinforcement learning policy was formulated by combining the output impedance and desired trajectory, and a reward term was designed to implement the learned variable impedance policy. 
In \cite{yang2018dmps}, dynamic movement primitives (DMP) were introduced to learn and reproduce the movement trajectory and stiffness profile learned from demonstrations.
In \cite{jin2022optimal}, a Gaussian-mixture-model-based, state-dependent variable impedance method was proposed to transfer human impedance behavior to the robot while ensuring stability.

Despite significant advancements in VIC for exoskeleton robots, most of the existing control schemes cannot detect or alleviate physical conflicts, e.g., motion asynchronization, human fatigue, and imbalance during walking, resulting in potential safety issues.

\section{Preliminaries}
\label{preli}
\begin{figure}[!h]
    \centering
    \includegraphics[width=4.0cm]{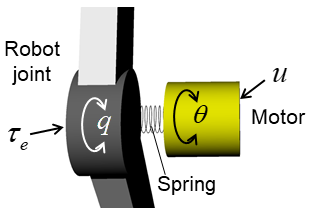}
    \caption{Illustration of a robot joint driven by a series elastic actuator (SEA), where a spring is installed between the driving motor and robot joint. Specifically, $\bm u$ is the control input exerted by the motor, $\bm\tau_e$ denotes the interaction torque, $\bm q$ is the rotational angle of the robot joint, and $\bm\theta$ denotes the rotation of the motor rotor.}
    \label{FigSEA}
\end{figure}

The exoskeleton robot considered in this study is driven by series elastic actuators (SEAs), the principle of which is illustrated in Fig.\ref{FigSEA}. 
Unlike rigid actuators, an elastic element (e.g., a spring) is installed between the driving motor and robot joint, which can store excessive energy and tolerate impacts, thereby ensuring structural safety.

The dynamic model of an SEA-driven exoskeleton robot can be described as:
\begin{eqnarray}
&\bm M(\bm q)\ddot{\bm q}+\bm C(\dot{\bm q}, \bm q)\dot{\bm q}+\bm g(\bm q)=\bm K(\bm\theta-\bm q)+\bm\tau_e,\label{dynRobot}\\
&\bm B\ddot{\bm\theta}+\bm K(\bm\theta-\bm q)=\bm u, \label{dynSEA}
\end{eqnarray}
where (\ref{dynRobot}) and (\ref{dynSEA}) describe the subsystems at the robot side and the actuator side, respectively, which are coupled with the output torque of the SEA, that is, $\bm\tau_o=\bm K(\bm\theta - \bm q)$. Specifically, $\bm q\hspace{-0.05cm}\in\hspace{-0.05cm}\Re^n$ denotes the vector of the joint angles of the robot; $n$ is the number of degrees of freedom (DoFs); $\bm \theta\hspace{-0.05cm}\in\hspace{-0.05cm}\Re^n$ denotes the vector of the rotational angles of the rotors; $\bm K\hspace{-0.05cm}\in\hspace{-0.05cm}\Re^{n\times n}$ is the stiffness matrix; $\bm M(\bm q)$ and $\bm B\hspace{-0.05cm}\in\hspace{-0.05cm}\Re^{n\times n}$ are the inertia and damping matrices for both the robot and actuator, respectively; $\bm C(\dot{\bm q}, \bm q)\hspace{-0.05cm}\in\hspace{-0.05cm}\Re^{n\times n}$ is a matrix related to the centripetal and Coriolis forces; $\bm g(\bm q)\hspace{-0.05cm}\in\hspace{-0.05cm}\Re^{n}$ is a vector of gravitational torque; $\bm\tau_e\hspace{-0.05cm}\in\hspace{-0.05cm}\Re^n$ is a vector related to physical interaction; and $\bm u\hspace{-0.05cm}\in\hspace{-0.05cm}\Re^n$ denotes the control input, which is also the torque exerted by the driving motor. 

The dynamic model described by (\ref{dynRobot}) and (\ref{dynSEA}) exhibits the following properties
\cite{cheah2015task,arimoto1996control}:
\begin{enumerate}
\item[1)] The matrix $\bm M(\bm q)$ is bounded, symmetric, and positive definite;

\item[2)] The term $\dot{\bm M}(\bm q)-2\bm C(\bm q, \dot{\bm q})$ is skew-symmetric;

\item[3)] The matrices $\bm B$ and $\bm K$ are constant, diagonal, and positive definite.

\end{enumerate}

We assume that the dynamic parameters (i.e., $\bm M(\bm q)$, $\bm C(\dot{\bm q}, \bm q)$, $\bm g(\bm q)$, $\bm K$, $\bm B$) are known or can be obtained with sufficient accuracy from datasheets or using calibration and identification techniques \cite{olsen2001new}.
\begin{figure*}[!t]
    \centering
    \includegraphics[width=15cm]{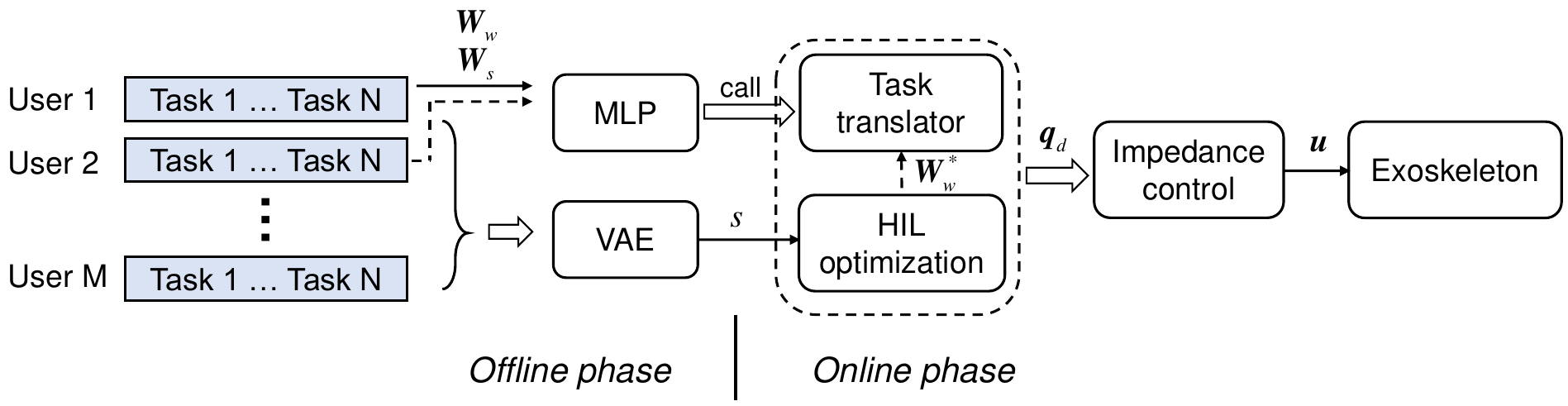}
    \caption{Workflow of the proposed learning and control method, where $M$ is the number of human users, $N$ denotes the number of tasks, $\bm W_w$ and $\bm W_s$ denote the learned parameters from different tasks, $s$ is the anomaly score, $\bm W_w^{*}$ represents the optimized parameters for a specific human user, $\bm q_d$ is the generated trajectory, and $\bm u$ is the control input.}
    \label{overall_frame}
    \vspace{-0.3cm}
\end{figure*}

The dynamic model of the overall system, described by (\ref{dynRobot}) and (\ref{dynSEA}), exhibits different time scales. Specifically, the SEA subsystem is slower, and the robot subsystem is faster. To control this system according to the singular perturbation theory \cite{spong1987modeling}, the control input can be designed as
\begin{eqnarray}
&\bm u=\bm u_f +\bm u_s,\label{controlSection3}
\end{eqnarray}
where $\bm u_f$ denotes the fast time-scale control term for stabilizing 
(\ref{dynSEA}), and $\bm u_s$ denotes the slow time-scale control term to stabilize 
(\ref{dynRobot}). 
A representative $\bm u_f$ can be expressed as
\begin{eqnarray}
&\bm u_f=-\bm K_v(\dot{\bm\theta}-\dot{\bm q}),\label{uf}
\end{eqnarray}
where $\bm K_v\in\Re^{n\times n}$ is a diagonal and positive-definite matrix. 

Substituting (\ref{controlSection3}) and (\ref{uf}) into (\ref{dynSEA}) yields
\begin{eqnarray}
\bm B\ddot{\bm\theta}+\bm K(\bm\theta-\bm q)+\bm K_v(\dot{\bm\theta}-\dot{\bm q})=\bm u_s,
\end{eqnarray}
which can be rewritten as
\begin{eqnarray}
\bm B(\ddot{\bm\theta}-\ddot{\bm q})+\bm K(\bm\theta-\bm q)+\bm K_v(\dot{\bm\theta}-\dot{\bm q})=\bm u_s-\bm B \ddot{\bm q}.
\label{substitute_SP1sec}
\end{eqnarray}

Note that $\bm \tau_o=\bm K(\bm\theta-\bm q)$. By introducing $\bm K = \bm K_1/\varepsilon ^{2}$ and $\bm K_v = \bm K_2/\varepsilon $, with $\varepsilon$ being a small positive parameter, (\ref{substitute_SP1sec}) can be written as
\begin{eqnarray}
&\varepsilon^2\bm B\ddot{\bm \tau}_o+\varepsilon \bm K_2\dot{\bm \tau}_o+\bm K_1\bm \tau_o=\bm K_1(\bm u_s-\bm B \ddot{\bm q}).
\label{substitute_SP3sec}
\end{eqnarray}
When $\varepsilon \hspace{-0.05cm}=\hspace{-0.05cm}0$, the solution of (\ref{substitute_SP3sec}) is $\Bar{\bm \tau}_o=\bm u_s-\bm B \ddot{\bm q}$.

If the fast time-scale is set as $\gamma=\frac{t}{\varepsilon }$, $\Bar{\bm \tau}_o$ is achieved at $\gamma\rightarrow\infty$. Note that $\Bar{\bm \tau}_o$ remains constant at $\varepsilon =0$.
Thus, a new variable is introduced as $\bm \eta=\bm \tau_o-\Bar{\bm \tau}_o$ to rewrite (\ref{substitute_SP3sec}) on the fast time-scale:
\begin{eqnarray}
\bm B(\frac{\rm d^2\bm \eta}{\rm d\gamma^2})+\bm K_2(\frac{\rm d\bm \eta}{\rm d\gamma})+\bm K_1\bm \eta = \bm 0
\label{boundary_Layer},
\end{eqnarray}
which is referred to as the \textit{boundary-layer system}.

Then, substituting this solution into (\ref{dynRobot}), a \textit{quasi-steady-state system} can be derived as
\begin{eqnarray}
&(\bm M(\bm q)+\bm B)\ddot{\bm q}+\bm C(\dot{\bm q}, \bm q)\dot{\bm q}+\bm g(\bm q)=\bm u_s+\bm\tau_e.
\label{substitute_SP4sec}
\end{eqnarray}
According to the singular perturbation theory, the stability of the overall system is guaranteed if both the \textit{boundary-layer system} and the \textit{quasi-steady-state system} are exponentially stable. 



In addition, the control objective for (\ref{substitute_SP4sec}) can be  specified as
\begin{eqnarray}
&\bm C_d(\dot{\bm q}-\dot{\bm q}_d)+\bm K_d(\bm q-\bm q_d)=\bm\tau_e,\label{impedancemodel}
\end{eqnarray}
which describes a first-order impedance model with the desired damping and stiffness matrices (i.e., $\bm C_d, \bm K_d\in\Re^{n\times n}$), allowing the robot joint to deviate from the reference trajectory $\bm q_d\in\Re^n$, thereby ensuring safe interaction. Note that the desired inertia is neglected, matching the lower-limb model of human users \cite{oatis1993use}. 
In summary, the wearer's safety is ensured by both hardware (i.e., SEA) and software (e.g., impedance control) considerations in the proposed framework.

\section{Overall workflow}
The workflow of the proposed learning and control framework is shown in Fig. \ref{overall_frame}.
The individualized trajectory is generated through the offline learning phase and then HIL optimization is performed. 
In the offline stage, multiple human users wear the exoskeleton to perform various tasks (e.g., walking on the ground or ascending stairs, etc.). The joint trajectories of the exoskeleton robot are recorded and represented as the joint motion of human subjects. During the collection of datasets, the robot passively follows the subject under a zero-impedance model \cite{zimmermann2020towards} to simulate the user's natural movements without the exoskeleton. 

The collected datasets are used to train anomaly detection and task translation modules. 
The anomaly detection module shapes the impedance parameter (with the weighting function) and guides HIL optimization, while task translation establishes the mappings among multiple tasks.

In the online phase, HIL optimization is performed to customize the gait trajectory for a human user, considering the body parameters and walking habits among other factors.
This trajectory can be adapted for new tasks (e.g., ascending stairs), using the trained task translator.
The trajectory is integrated into a variable impedance model, and the robot tracks this model to provide safe and effective assistance.
Both HIL optimization and impedance control operate simultaneously in the online phase. 

The proposed framework offers several advantages, including ease of implementation and adaptability to multiple tasks and subjects:
\begin{enumerate}
    \item[-] Only proprioceptive sensors, which are typically integrated into the exoskeleton, are required for trajectory learning and interaction control, resulting in cost effectiveness and reduced complexity;
    \item[-] Using common sensors, the robot can effectively adapt to new users and tasks that are not included in the offline training;
    \item[-] This adaptation process requires the data of only a few users to fit the nonlinear relationships between different patterns.
\end{enumerate}

\begin{figure*}[!t]
    \centering
    \includegraphics[width=17cm]{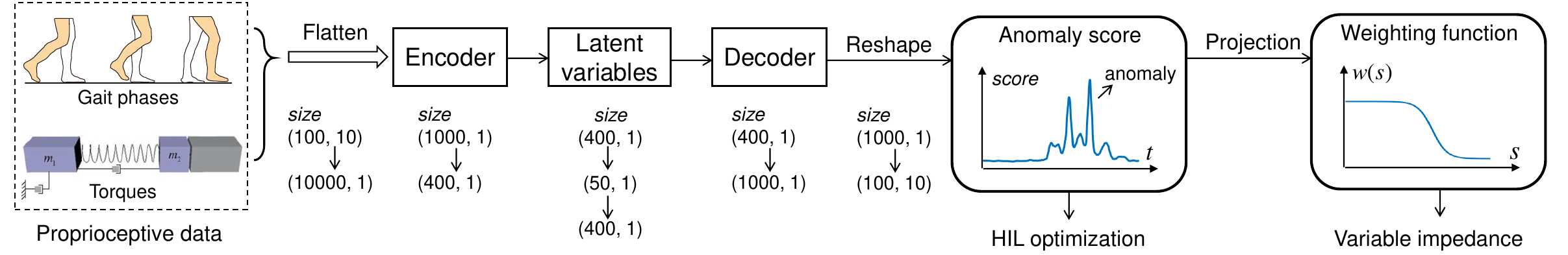}
    \caption{An illustration of the anomaly detection structure, where the proprioceptive sensors are utilized to generate the anomaly score, which is projected into a weighting function for variable impedance control.}
    \label{VAEnetwork}
    \vspace{-0.3cm}
\end{figure*}

\section{Offline Learning}
\label{offline_sec}
As shown in Fig. \ref{overall_frame}, the dataset of multiple human users in multiple tasks is used to train the anomaly detection and task translation modules. \\

\noindent\textbf{Anomaly Detection}: During closely coupled interaction, a mismatch between the
wearer and robot may result in physical conflict, which
can adversely influence the assistance efficiency and safety.
The proposed framework detects and then alleviates such conflicts.

The anomaly detection module is based on a VAE (Fig. \ref{VAEnetwork}).
The VAE, receives two types of information from proprioceptive sensors (i.e., encoders and force sensors built into the exoskeleton): interaction torques (based on spring deflection measurements) and gait phases (based on rotation angle measurements) and outputs a continuous anomaly score. 



To handle time-series signals, a sliding window is introduced to segment the data, which are then input to the network.
In the proposed network, latent variables are used to capture normal patterns of the input signals and then reconstruct the original data.
The anomaly score is obtained by calculating the reconstruction error, 
which increases when the physical conflict is significant and decreases when the conflict is relaxed.

To train the proposed network, the latent variable $\rho$ is re-parameterized as \cite{kingma2013auto}
\begin{eqnarray}
&\rho=\mu+\sigma \odot \epsilon,
\end{eqnarray}
where $\mu$ and $\sigma$ are the mean and standard deviation obtained from the encoder, respectively. Sample noise is set to obey the standard normal distribution i.e., $\epsilon \sim \mathcal{N}(0,\bm I)$.

The loss function can be formulated as
\begin{eqnarray}
&\mathcal{L}_{VAE}(\mathbf{x}^t)={||\mathbf{x}^t- \hat{\mathbf{x}}^t||}^{2}+KL[\mathcal{N}(\mu,\sigma),\mathcal{N}(0,\bm I)],
\end{eqnarray}
which consists of both the reconstruction error and the Kullback–Leibler-divergence,
with $\hat{\mathbf{x}}^t$ denoting the output of the decoder. 
Then, the anomaly score can be defined as
\begin{eqnarray}
&s= MSE(\mathbf{x}^t, \hat{\mathbf{x}}^t)={||\mathbf{x}^t- \hat{\mathbf{x}}^t||}^{2}.
\end{eqnarray}
Training is performed by inputting the data of $M$ human users, with $N$ tasks for each user, into the constructed VAE. This anomaly detection model is trained in an unsupervised manner, and its output score represents the conflict among different wearers and tasks.\\

\noindent\textbf{Task Translation}:
All collected datasets are used to train the task translator module, to uncover the relationships among multiple tasks. 
The proposed task translator is based on a Multi-Layer Perceptron (MLP) \cite{li2021bear}. 
Unlike VAE-based training, each entry fed into the task translator represents parameterized trajectories (i.e., weight matrices) for various tasks performed by a single user.
Details of the trajectory encoding are presented in Section \ref{HIL_opt_sec}. The identified relationships allow the robot to transfer learned tasks to new ones in the subsequent online HIL phase.
These tasks are typically not repeatable (e.g., ascending stairs and standing up) and thus impractical for online learning. 
\begin{figure}[!ht]
    \centering
    \includegraphics[width=8.8cm]{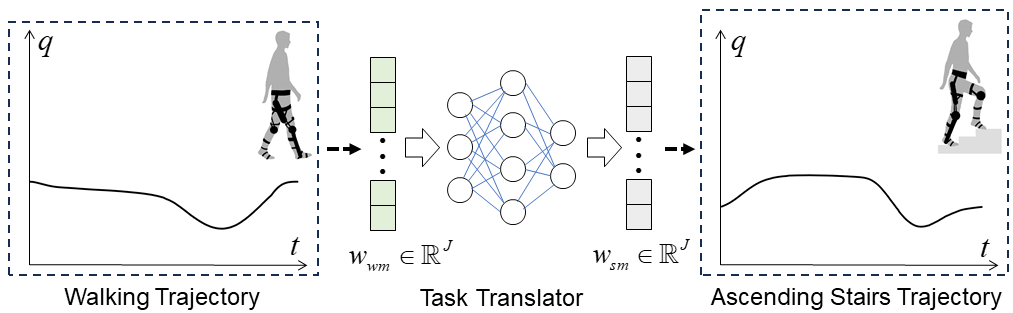}
    \caption{The task translator can generate the trajectory for a new task that is impractical to learn online (e.g., ascending stairs).
    Here, $\bm w_{wm}$ and $\bm w_{sm}$ represent the encoded weight vectors for user $m$ during walking and stair climbing tasks, respectively. }
    \label{trans_ws}
\end{figure}

Fig. \ref{trans_ws} illustrates the workflow of task translation, where $\bm w_{wm}$ and $\bm w_{sm}$ represent the encoded weight vectors for user $m$ during the tasks of walking and ascending stairs, respectively. These vectors are processed through DMP, as discussed in Section \ref{HIL_opt_sec}.
To train the task translator, gait trajectories in the different tasks are collected in weight matrix form as a training set $(\bm W_{wm},\bm W_{sm})$, 
where $\bm W_{wm}$ and $\bm W_{sm}$ are the weight matrices of user $m$ during the tasks of walking and ascending stairs, respectively.
Then, the MLP is used to determine the mapping relationship of the weight matrices under different tasks.
The predictive weight matrix in the new task is obtained as
\begin{align}
    &\hat{\bm W}_{sm} = \bm F_r(\bm W_{wm}),
\end{align}
where $\bm F_r$ 
is a nonlinear regression function yielded by a neural network.

The following loss function is minimized
\begin{align}
    & \mathcal{L} = \sum_{m}^{M}l({\bm W}_{sm},\hat{\bm W}_{sm}),
\end{align}
where $l$ is a differentiable convex loss function that represents the distance between the prediction $\hat{\bm W}_{sm}$ and target ${\bm W}_{sm}$, and $M$ is the number of users. 
Function $l$ is written in the following quadratic form:
\begin{align}
    &l(\bm W, \hat{\bm W}) = \Vert \bm W - \hat{\bm W} \Vert_{F}^{2},
    \label{littile_loss_function}
\end{align}
where $\Vert \cdot \Vert_{F}$ indicates the Frobenius norm.

Compared with techniques commonly used for prediction, such as K-Nearest Neighbors (KNN), Gaussian Process Regression (GPR), Extreme Gradient Boosting (XGBoost), Gradient Boosting (GBoost), and Support Vector Machines (SVM), the proposed NN-based task translator exhibits better generalization performance over a limited dataset, as validated experimentally.

Notably, the weight matrix is an intrinsic property of the user, which implicitly includes parameters affecting the gait trajectory, such as the height and arm length \cite{li2021bear}.
For trajectory recording, the proposed task translator requires only proprioceptive sensors.

\section{HIL Optimization}
\label{HIL_opt_sec}
\begin{figure*}[!t]
    \centering
    \includegraphics[width=13cm]{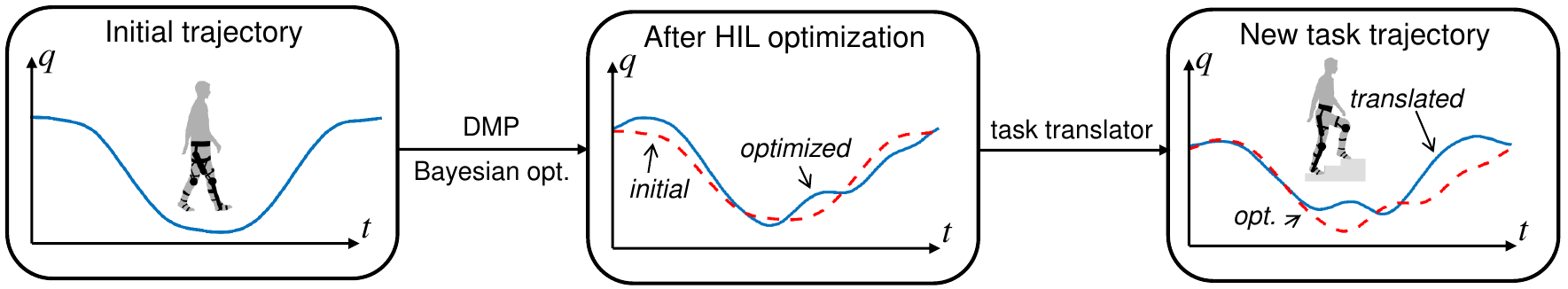}
    \caption{An illustration of the knee joint's trajectory within one gait cycle, where the three figures show the initial trajectory, the individualized trajectory for a specific subject, and the trajectory for a new task respectively.}
    \label{learning_for_wearer_frame}
    \vspace{-0.3cm}
\end{figure*}

This section presents a new HIL optimization scheme (Fig. \ref{learning_for_wearer_frame}) for exoskeleton robots to facilitate their online adaptation to different wearers (with various walking habits and body parameters) across multiple tasks. 
This adaptability is important for robots to exhibit operational flexibility in real-world applications. 

First, the proposed scheme parameterizes the gait trajectory of the user based on DMP \cite{ijspeert2013dynamical}. 
%
Specifically, the following \textit{output system} is constructed to encode the desired trajectory:
\begin{align}
& \dot \omega_{i} = -\Omega_o\left[\alpha_o(\beta_o q_{di} + \omega_i)+\frac{\sum_{j=1}^{J}\Psi_j w_{ij}}{\sum_{j=1}^{J}\Psi_j}\right],\\
& \dot q_{di} = \Omega_o \omega_{i},\\
& \Psi_j = \exp [h(\cos(\Omega_o t-c_j)-1)],
\end{align}
where $\Omega_o=2\pi$ is the fundamental frequency in a normalized cycle, $w_{ij}$ represents the weights of the $i$-th joint to be determined, $q_{di}$ denotes the $i$-th joint's desired trajectory normalized in the time scale, 
and $\Psi_j$ represent Gaussian-like kernel functions with a width $h$ and center $c_j$. In practice, after determining the number of kernels $J$, $h$ is typically set as $2.5J$ and $c_j$ is evenly spaced in $[0,2\pi]$. 

For the sake of brevity, the parameterized trajectory can be written as
\begin{align}
    &\bm q_d = f_p(t,\bm W, \Omega_o),
    \label{overall_output_system}
\end{align}
where $\bm W$ is the weight matrix with elements $w_{ij}$ and $f_p$ is the function governing the \textit{output system} to produce the joint trajectory $\bm q_d$. 
Another
\textit{canonical system} is implemented to extract the motion frequency $\Omega_i$ of each joint
\begin{align}
&\dot \phi_{ij} = j\Omega_i - k_{c1} e_i \sin(\phi_{ij}),\\
&\dot \Omega_i = -k_{c1} e_i \sin(\phi_{i1}),\\
\label{canonical_ome}
&\Tilde{q}_{di} =  \sum_j A_{ij} \cos(\phi_{ij}),\\
&\dot A_{ij} =  k_{c2}e_i\cos(\phi_{ij}),\\
&e_i = q_{i} - \Tilde{q}_{di},
\end{align}
where $q_{i}$ is the profile of the $i$-th joint; $\phi_{ij}$ and $A_{ij}$ represent the phase and amplitude of the $j$-th component in the $i$-th joint, respectively; and $k_{c1}$ and $k_{c2}$ are positive constants. 

Discrete movements \cite{ijspeert2013dynamical}, such as sitting and standing, can be represented using DMP within a unified framework by encoding them as a weight matrix $\bm W$. Additionally, because the time required to perform the movement does not vary significantly across individuals, speed modulation is not required.
In a scenario with $N_a$ active joints, the overall assistance can be expressed as
\begin{align}
    & \Omega_c = \frac{1}{N_a}\sum_{i}^{N_a} \Omega_i,\label{eq:omega}\\
    &\bm q_d = \left\{ 
    \begin{array}{*{20}{l}}
     f_p(t,\bm W, \Omega_c),&& \textit{periodic movement}\\
     f_d(t,\bm W),&& \textit{discrete movement}
    \end{array} 
    \right.
    \label{overall_output_system}
\end{align}
where $f_d(t,\bm W)$ is the second-order dynamic system under discrete movement conditions.

Next, a comprehensive cost function is formulated using both the parameterized trajectory and anomaly score as
\begin{align}
 \mathcal{J}_{HIL}(s,\bm q,\bm W) = & \frac{\lambda}{T}\sum_{i}^{N_a} \sum_{j}^{T}( q_{di}^{(j)} - q_{i}^{(j)})^2+\frac{1-\lambda}{T}\sum_{j}^{T} s^{(j)},
 \label{cost_function}
\end{align}
where the superscript and subscript indicate the time index and joint index in a gait cycle, respectively. The first term ensures that the generated trajectory matches the specific user, and the second term helps ignore the abnormal patterns of the user (likely attributable to physical conflicts) during the learning process.
The weight of these two terms is governed by the hyper-parameter 
$\lambda \hspace{-0.05cm}\in \hspace{-0.05cm}(0,1)$. 
Specifically, when $\lambda = 0$, optimization is based on only the collected dataset. 
Conversely, when $\lambda = 1$, the emphasis shifts to optimizing the trajectory alignment.

Bayesian optimization is applied to determine the global minimum of $\mathcal{J}_{HIL}(s,\bm q,\bm W)$ under a bounded feasible set, which has proven effective in previous HIL protocols \cite{kim2019bayesian}.
To this end, the selected matrices and their associated costs are collated into $\mathcal{W}_{past}$ and $\bm{\mathcal{J}}_{past}$ respectively. 
Note that $\bm{\mathcal{J}}_{past}$ is formulated as a vector.
Subsequently, we model the target function using Gaussian processes as
\begin{align}
&\left[ 
\begin{array}{c}
\mathcal{J}_{new} \\
\bm{\mathcal{J}}_{past} \\
\end{array} 
\right]\sim \mathcal{N}\left( \bm 0,
\left[ 
\begin{array}{cc}
K_{nn} & \bm K_{pn}^T \\
\bm K_{pn} & \bm K_{pp} \\
\end{array} 
\right]
\right),
\end{align}
where $\mathcal{J}_{new}$ represents the cost corresponding to the weight matrix $\bm W^{new}$ to be identified.
Notably, setting the mean as zero simplifies the calculation, which has proven effective in practice \cite{kim2019bayesian} as most of the information is contained in the covariance.
Using the radial basis function as the kernel function, $k(\bm x,\bm x' )$, we introduce kernel matrices $K_{nn},\bm K_{pn},$, and $\bm K_{pp}$ with the following elements:
\begin{align}
&K_{nn} = k(Vec(\bm W^{new}),Vec(\bm W^{new}) ),\\
&K_{in} = k(Vec(\bm W_i),Vec(\bm W^{new}) ),\quad \bm W_i\in \mathcal{W}_{past}\\
&K_{ij} = k(Vec(\bm W_i),Vec(\bm W_j) ). \quad\bm W_i,\bm W_j\in \mathcal{W}_{past}
\end{align}
Here, $Vec(\cdot)$ denotes a vectorization function.
Following the construction of the posterior distribution, we use the Lower Confidence Bound (LCB) as the acquisition function for guiding the selection of the subsequent point for exploration:
\begin{align}
\label{W_new}
&\bm W^{new} = \underset{\bm W^{new}}{\operatorname{argmin}} \mu_{new} - \upsilon\sigma_{new},\\
& \mu_{new} = \bm K_{pn}^T\bm K_{pp}^{-1}\bm{\mathcal{J}}_{past},\\
&\sigma_{new} = K_{nn} - \bm K_{pn}^T\bm K_{pp}^{-1}\bm K_{pn},
\end{align}
where $\upsilon$ is a positive hyperparameter.
The minimization of (\ref{W_new}) is implemented using the L-BFGS algorithm \cite{liu1989limited}.
This process enables iterative updates of the posterior distribution based on historical data, and the appropriate weight matrix is then derived from the acquisition function. 
To accelerate the convergence over the extended dimensions compared with previous research, the initial parameters are specified based on the sampled gait.
Subsequently, the optimal weight matrix $\bm W^*$ for the walking task can be learned based on the proposed cost function.

Minimizing (\ref{cost_function}) can yield an individualized and anomaly-robust trajectory (i.e., $\bm q_d$), allowing the robot to learn the correct and appropriate gait trajectory under all conditions to improve the effectiveness of assistance and the wearer’s comfort.

After the walking optimization $\bm W_{wm}^{*}$ for user $m$ is obtained, the optimal weight matrix for a new task can be predicted using the proposed task translator as
\begin{align}
    &\hat{\bm W}_{sm}^{*} = \bm F_r(\bm W_{wm}^{*}), 
\end{align}
which enables the provision of predictive assistance $\hat{\bm W}_{sm}^{*}$ without the demonstration of a new task based on the known optimization result.
Thus, the proposed HIL optimization framework can effectively adapt to multiple wearers and tasks.


\section{Variable Impedance Control}
The individualized and generalized trajectory is embedded into a variable impedance model to regulate the robot's action to provide safe and effective assistance. The impedance model is varied according to the anomaly score presented in Section \ref{offline_sec} to alleviate physical conflicts with the wearer. We consider the following aspects:
\begin{enumerate}
    \item[-] When the anomaly score oscillates around a small value, the robot should ignore such minor conflicts to avoid over-reaction;
    
    \item[-] When the anomaly score exceeds a certain threshold, significant conflicts may arise, which must be promptly addressed by the robot.
\end{enumerate}

To this end, the anomaly score is mapped into a weighting function as
\begin{eqnarray}
&w(s) = \lambda_1 \tanh(-\frac{s}{\chi_1} + \chi_2) +\lambda_2, \label{weight_fun}
\end{eqnarray}
where $\lambda_1$ and $\lambda_2$ are positive constants that determine the range and median of the weighting function, respectively;
$\chi_1$ is a constant that normalizes the anomaly score into a specified small range; and $\chi_2$ represents the offset of the weighting function from the origin of the coordinates along the positive horizontal axis. 

Next, the desired impedance model (\ref{impedancemodel}) is reformulated with the weighting function as
\begin{eqnarray}
&\bm C_d(\dot{\bm q}-\dot{\bm q}_d)+\bm K_d(\bm q-\bm q_d)=\frac{1}{w(s)}\bm\tau_e.\label{model}
\end{eqnarray}
Multiplying both sides of (\ref{model}) with $w(s)$ yields
\begin{eqnarray}
&\bm C_d(t)(\dot{\bm q}-\dot{\bm q}_d)+\bm K_d(t)(\bm q-\bm q_d)=\bm\tau_e,\label{model2}
\end{eqnarray}
where $\bm C_d(t)\stackrel{\triangle}{=}w(s)\bm C_d$ and $\bm K_d(t)\stackrel{\triangle}{=}w(s)\bm K_d$, varying according to the weighting function. Specifically, when the anomaly score increases, $w(s)$ is reduced, thus lowering the impedance parameters and ensuring that the robot reacts passively to the conflict. Otherwise, the impedance parameters return to their original values to maintain the assistance. 

Then, an impedance vector is introduced as
\begin{align}
\bm z&=\dot{\bm q}-\dot{\bm q}_r\nonumber \notag \\
&=\dot{\bm q}-\dot{\bm q}_d+\bm C_d^{-1}\bm K_d(\bm q-\bm q_d)-\frac{1}{w(s)}\bm C_d^{-1}\bm\tau_e,\label{vectorz}
\end{align}
where 
\begin{eqnarray}
&\dot{\bm q}_r=\dot{\bm q}_d-\bm C_d^{-1}\bm K_d(\bm q-\bm q_d)+\frac{1}{w(s)}\bm C_d^{-1}\bm\tau_e
\end{eqnarray}
is a reference vector. According to (\ref{vectorz}), the convergence of $\bm z\rightarrow\bm 0$ implies the realization of the desired impedance model (\ref{model}).

Notably, the impedance vector (\ref{vectorz}) depends on the interaction torque $\bm\tau_e$. 
To eliminate the use of sensors for torque measurement, a model-based disturbance observer (DOB) \cite{mohammadi2017nonlinear} is established to estimate the interaction torque:
\begin{eqnarray}
    \left \{
    \begin{array}{*{20}{l}}
    \dot{\bm y} = -\bm L(\dot{\bm q}, \bm q){\bm y}-\bm L(\dot{\bm q}, \bm q)[\bm K(\bm\theta-\bm q)\\
\qquad\quad-\bm C(\dot{\bm q}, \bm q)\dot{\bm q}-\bm g(\bm q)+\bm p(\dot{\bm q}, \bm q)]
    \\
    \hat{\bm\tau}_e=\bm y+\bm p(\dot{\bm q}, \bm q)
    \end{array} \right.,
\label{dob_realization}
\end{eqnarray}
where $\hat{\bm\tau}_e\in\Re^n$ denotes the estimated interaction torque, and $\bm y \in\Re^n$ is an auxiliary vector used to make the observer independent of acceleration information. 

According to \cite{mohammadi2013nonlinear}, the observer gain matrix $\bm L(\dot{\bm q}, \bm q)\in\Re^{n\times n}$ and vector $\bm p(\dot{\bm q}, \bm q)\in\Re^n$ can be expressed as
\begin{eqnarray}
    \left \{
    \begin{array}{*{20}{l}}
    \bm L(\bm q,\dot{\bm q}) = \bm A^{-1}\bm M^{-1}(\bm q)
    \\
    \bm p(\dot{\bm q}, \bm q)=\bm A^{-1}\dot{\bm q}
    \end{array} \right.
\label{dob_gain}
\end{eqnarray}
where $\bm A\hspace{-0.05cm}\in\hspace{-0.05cm}\Re^{n\times n}$ denotes a constant and invertible matrix to be determined. 

With the definition of the observation error as $\Tilde{\bm\tau}_e=\hat{\bm\tau}_e-\bm\tau_e$, the closed-loop equation of the DOB can be given as
\begin{eqnarray}
&\dot{\Tilde{\bm\tau}}_e= -\bm L(\bm q,\dot{\bm q})\Tilde{{\bm\tau}}_e-\dot{\bm\tau}_e
\label{dob_error}.
\end{eqnarray}

The following proposition can be derived to guarantee the convergence of ${\Tilde{\bm\tau}}_e$.

\noindent\textbf{Proposition 1}:
{\em In the presence of $\Vert \dot{\bm\tau}_e \Vert \hspace{-0.05cm}\leq\hspace{-0.05cm} \zeta$ where $\zeta$ denotes the bound, the disturbance tracking error ${\Tilde{\bm\tau}}_e$ is globally uniformly ultimately bounded, when $\bm A$ is chosen such that the following condition is satisfied
\begin{eqnarray}
&\bm A + \bm A^T-\bm A^T \dot{\bm M}(\bm q)\bm A = \bm \Gamma > \bm 0
\label{A_require},
\end{eqnarray}
where $\bm \Gamma$ is a positive definite matrix.}

\begin{IEEEproof}
Consider the following candidate Lyapunov function:
\begin{eqnarray}
&V_e = \Tilde{\bm\tau}_e^T\bm A^T\bm M(\bm q)\bm A\Tilde{\bm\tau}_e
\label{Lya_tau}.
\end{eqnarray}
Differentiating $V_e$ with respect to time and substituting (\ref{dob_error}) into the expression yields
\begin{align}
\dot V_e = & -\Tilde{{\bm\tau}}_e^T(\bm A + \bm A^T - \bm A^T \dot{\bm M}(\bm q)\bm A)\Tilde{{\bm\tau}}_e \notag \\
& + \dot{\bm\tau}_e^T\bm A^T \bm M(\bm q)\bm A\Tilde{{\bm\tau}}_e + \Tilde{{\bm\tau}}_e^T\bm A^T \bm M(\bm q)\bm A \dot{\bm\tau}_e
\label{dot_Lya_tau}.
\end{align}

According to Schwartz inequality and the boundedness of $\Vert \dot{\bm\tau}_e \Vert$, it can be deduced that
\begin{align}
\dot V_e \leq& -\lambda_{\min}(\bm \Gamma)\Vert\Tilde{{\bm\tau}}_e \Vert^2 + 2\zeta\lambda_{\max}(\bm M(\bm q))\Vert\bm A \Vert^2 \Vert\Tilde{{\bm\tau}}_e \Vert \notag \\
=& -(1-\varrho)\lambda_{\min}(\bm \Gamma)\Vert\Tilde{{\bm\tau}}_e \Vert^2 - \varrho\lambda_{\min}(\bm \Gamma)\Vert\Tilde{{\bm\tau}}_e \Vert^2 \notag \\
& + 2\zeta\lambda_{\max}(\bm M(\bm q))\Vert\bm A \Vert^2 \Vert\Tilde{{\bm\tau}}_e \Vert
\label{dot_Lya_tau1},
\end{align}
where $\varrho \in (0,1)$. 

Because $V_e \leq \lambda_{\max}(\bm M(\bm q))\Vert\bm A \Vert^2\Vert\Tilde{{\bm\tau}}_e \Vert^2$, the inequality of
\begin{align}
\dot V_e \leq& -\frac{(1-\varrho)\lambda_{\min}(\bm \Gamma)}{\lambda_{\max}(\bm M(\bm q))\Vert\bm A \Vert^2} V_e
\label{dot_Lya_tau3}
\end{align}
is obtained if
\begin{align}
\Vert\Tilde{{\bm\tau}}_e \Vert \geq &\frac{2\zeta\lambda_{\max}(\bm M(\bm q))\Vert\bm A \Vert^2}{\varrho\lambda_{\min}(\bm \Gamma)} 
\label{dot_Lya_tau2}.
\end{align}

When (\ref{dot_Lya_tau2}) is satisfied, the last two terms of (\ref{dot_Lya_tau1}) are positive, and thus, inequality (\ref{dot_Lya_tau3}) is guaranteed. 

Therefore, $\Tilde{{\bm\tau}}_e$ converges asymptotically to the ball with radius $2\zeta\lambda_{\max}(\bm M(\bm q))\Vert\bm A \Vert^2/{\varrho\lambda_{\min}(\bm \Gamma)}$\cite{arimoto1996control}.
\end{IEEEproof}
The analytical solution of matrix $\bm A$ can be expressed as\cite{mohammadi2013nonlinear}

\begin{align}
    & \bm A^{-1} = \frac{1}{2} (\sigma_1 + 2\beta\sigma_2)\bm I,\\
    & \Vert\dot{\bm M}(\bm q)\Vert \leq \sigma_1, \Vert\bm M(\bm q)\Vert \leq \sigma_2, \notag
\end{align}
where $\beta$ is a positive constant related to the convergence rate. 


As mentioned in Section \ref{preli}, the overall control input is designed as (\ref{controlSection3}), with the fast time-scale control term defined as in (\ref{uf}). 
Next, using the estimated interaction torque $\hat{\bm\tau}_e$, the slow time-scale control term is established to stabilize (\ref{dynRobot}) and achieve the desired impedance model as
\begin{align}
    \bm u_s=&-\bm K_z\bm z-\hat{\bm\tau}_e-k_g\cdot {\bm{sgn}}(\bm z) \notag \\
    &+(\bm M(\bm q)+\bm B)\ddot{\bm q}_r+\bm C(\dot{\bm q}, \bm q)\dot{\bm q}_r
+\bm g(\bm q),\label{slow_controller}
\end{align}
where ${\bm{sgn}}(\cdot)$ is the sign function defined as
\begin{eqnarray}
{\rm sgn} (z) = 
\left\{ 
\begin{array}{*{20}{l}}
 1,&& z >0\\
 0,&& z= 0\\
 - 1,&& z<0
\end{array} 
\right.
\end{eqnarray}
where $k_g$ is a positive constant, and 
$\bm K_z\in\Re^{n\times n}$ is a diagonal and positive-definite matrix. 

The stability of the overall system is guaranteed if both the boundary-layer system
and quasi-steady-state system are exponentially stable. The following theorem can be stated:

\noindent\textbf{Theorem 1}:
{\em The proposed controller $\bm u$ in (\ref{controlSection3}) guarantees the stability of the overall system if $\bm K_1$ and $\bm K_2$ are positive definite and $k_g$ is adequately large.}

\begin{IEEEproof}
 First, the exponential stability of the \textit{boundary-layer system} in (\ref{boundary_Layer}) can be guaranteed by appropriately tuning $\bm K_1$ and $\bm K_2$. 

Then, 
substituting (\ref{vectorz}) and (\ref{slow_controller}) into (\ref{substitute_SP4sec}) yields
\begin{eqnarray}
(\bm M(\bm q)+\bm B)\dot{\bm z}+\bm C(\dot{\bm q}, \bm q)\bm z=-\bm K_z\bm z-\Tilde{\bm\tau}_e-k_g{\bm{sgn}}(\bm z).
\label{global_stability_pre}
\end{eqnarray}

Next, a Lyapunov-like candidate is proposed as
\begin{eqnarray}
&V = \frac{1}{2}\bm z^T(\bm M(\bm q)+\bm B)\bm z.
\label{Lyapunov_candidate}
\end{eqnarray}

Differentiating (\ref{Lyapunov_candidate}) with respect to time and substituting (\ref{global_stability_pre}) into it yields the following expression:
\begin{eqnarray}
\begin{array}{*{20}{l}}
\dot V = -\bm z^T\bm K_z\bm z - \bm z^T\Tilde{{\bm\tau}}_e-k_g\bm z^T\bm{sgn}(\bm z).
\end{array}
\label{diff_Lyapunov_candidate}
\end{eqnarray}

As shown in \textbf{Proposition 1}, when $\Vert \dot{\bm\tau}_e \Vert \hspace{-0.05cm} \leq \hspace{-0.05cm} \zeta$, the observation error is also bounded, and
\begin{align}
\dot V \leq& -\bm z^T\bm K_z\bm z -(k_g-\kappa)\Vert \bm z \Vert,
\end{align}
where
\begin{align}
\kappa \geq & \frac{2\zeta\lambda_{\max}(\bm M(\bm q))\Vert\bm A \Vert^2}{\varrho\lambda_{\min}(\bm \Gamma)} \notag
\label{diff_Lyapunov_candidate1}
\end{align}
is a constant, representing the bound of the observation error.

If $k_g$ is adequately large such that $k_g>\kappa$, we have
\begin{align}
\dot V \leq& -\bm z^T\bm K_z\bm z<0,
\end{align}

As $V>0$ and $\dot V < 0$, the \textit{quasi-steady-state system} is also exponentially stable. According to Tikhonov’s theorem \cite{tikhonov1952systems}, the stability of the closed-loop system is guaranteed, and convergence to the impedance vector is ensured.

\end{IEEEproof}

\section{Experiment}
\begin{figure}[!h]
    \centering
    \includegraphics[width=1\linewidth]{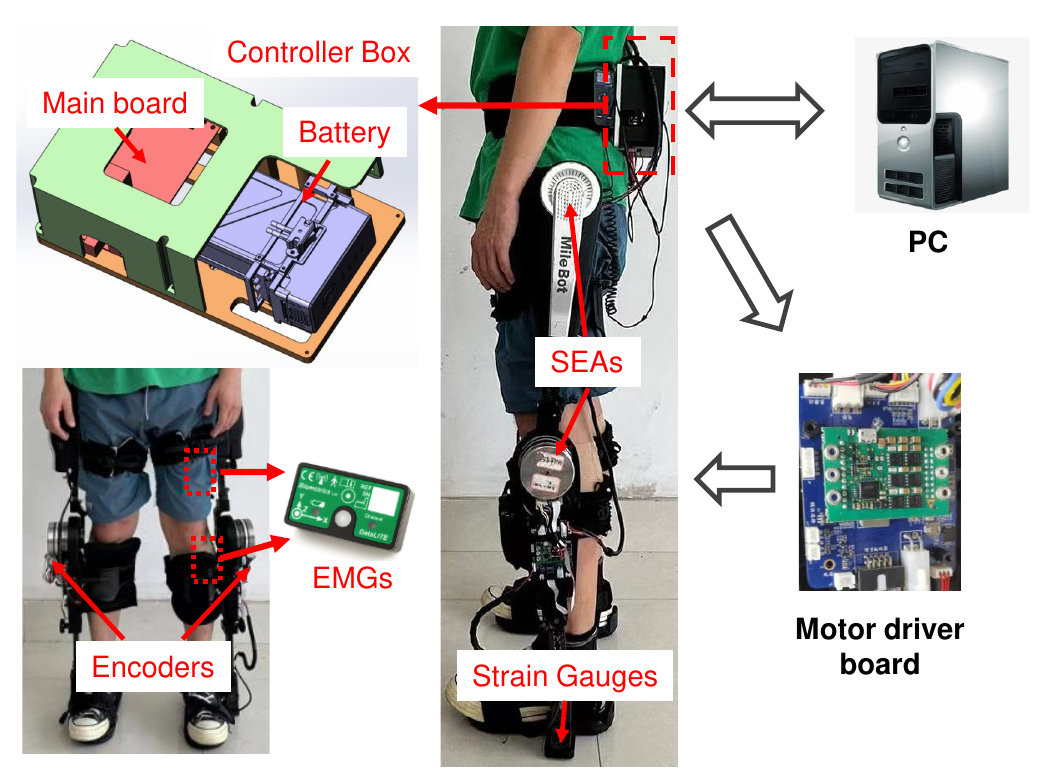}
    \caption{Experimental setup of the exoskeleton robot, where all four joints of both legs are driven by the SEAs. The controller box is composed of the main board and battery, combined with a PC to implement the detection network and control algorithm. The commands are transmitted to four slave boards to drive the robot joints.}
    \label{equipment}
\end{figure}

The proposed scheme was implemented in a bilateral lower-limb exoskeleton robot to assess its performance, as shown in Fig. \ref{equipment}. 
All four joints of the robot (two hip joints and two knee joints) were driven by SEAs.
The spring stiffness of each SEA was $635~Nm$, and the output torque was determined based on the spring deflection (measured using encoders) and known stiffness. 
The robot was equipped with two types of proprioceptive sensors: strain gauges and encoders (with 2048 lines). These sensors provided comprehensive task-specific information, including gait phases (measured using strain gauges), interaction torque, and movement trajectory (measured using encoders).
During the experiment, four wireless surface EMG sensors (Ws450, Biometrics Ltd.) were mounted on human leg muscles, specifically, the quadriceps femoris (QF) and tibialis anterior muscle (TA) of the left and right legs, to collect the feedback of human limb and hence assess the performance of the proposed controller.

\begin{figure*}[ht] 
	\centering 
	\vspace{-0.3cm} 
	\subfigtopskip=2pt 
	\subfigbottomskip=2pt 
	\subfigcapskip=-5pt 
	\subfigure[]{
		\label{walk_snap}
		\includegraphics[width=0.17\linewidth]{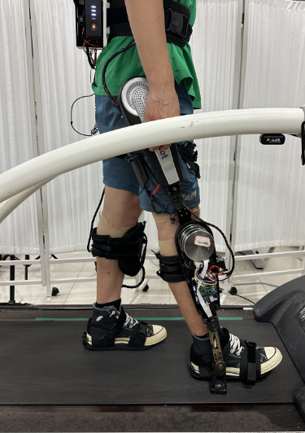}}
	\subfigure[]{
		\label{Up_snap}
		\includegraphics[width=0.17\linewidth]
        {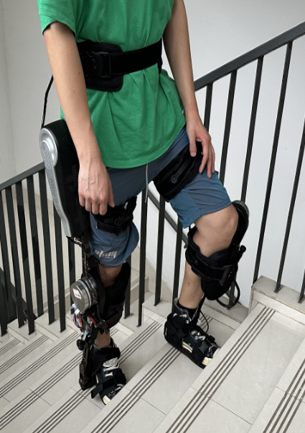}}
	\subfigure[]{
		\label{Down_snap}
		\includegraphics[width=0.17\linewidth]{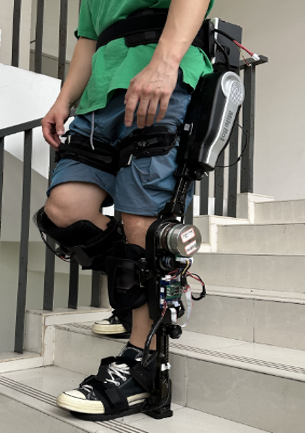}}
    \subfigure[]{
		\includegraphics[width=0.17\linewidth]{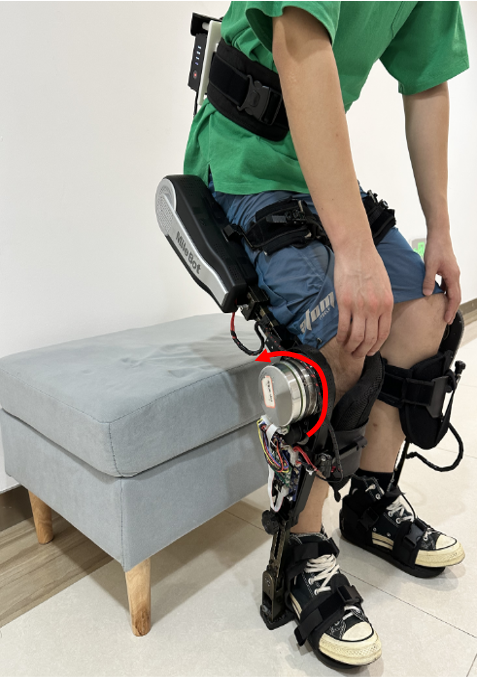}}
    \subfigure[]{
		\includegraphics[width=0.17\linewidth]{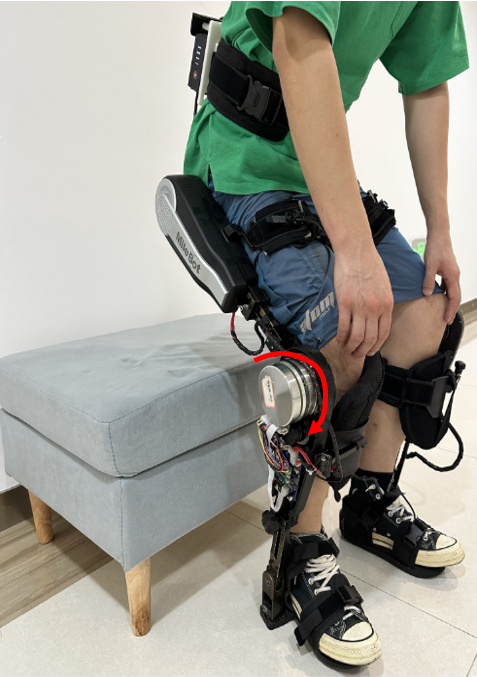}}
	\caption{Participants wearing the exoskeleton perform various tasks: (a) Walking; (b) Ascending stairs; (c) Descending stairs; (d) Squatting; (e) Standing up.}
	\vspace{0cm}
	\label{snapshot_task}
\end{figure*}

\begin{figure*}[ht] 
	\centering 
	\vspace{-0.3cm} 
	\subfigtopskip=2pt 
	\subfigbottomskip=2pt 
	\subfigcapskip=-5pt 
	\subfigure[]{
		\label{Slow_curve}
		\includegraphics[width=0.32\linewidth]{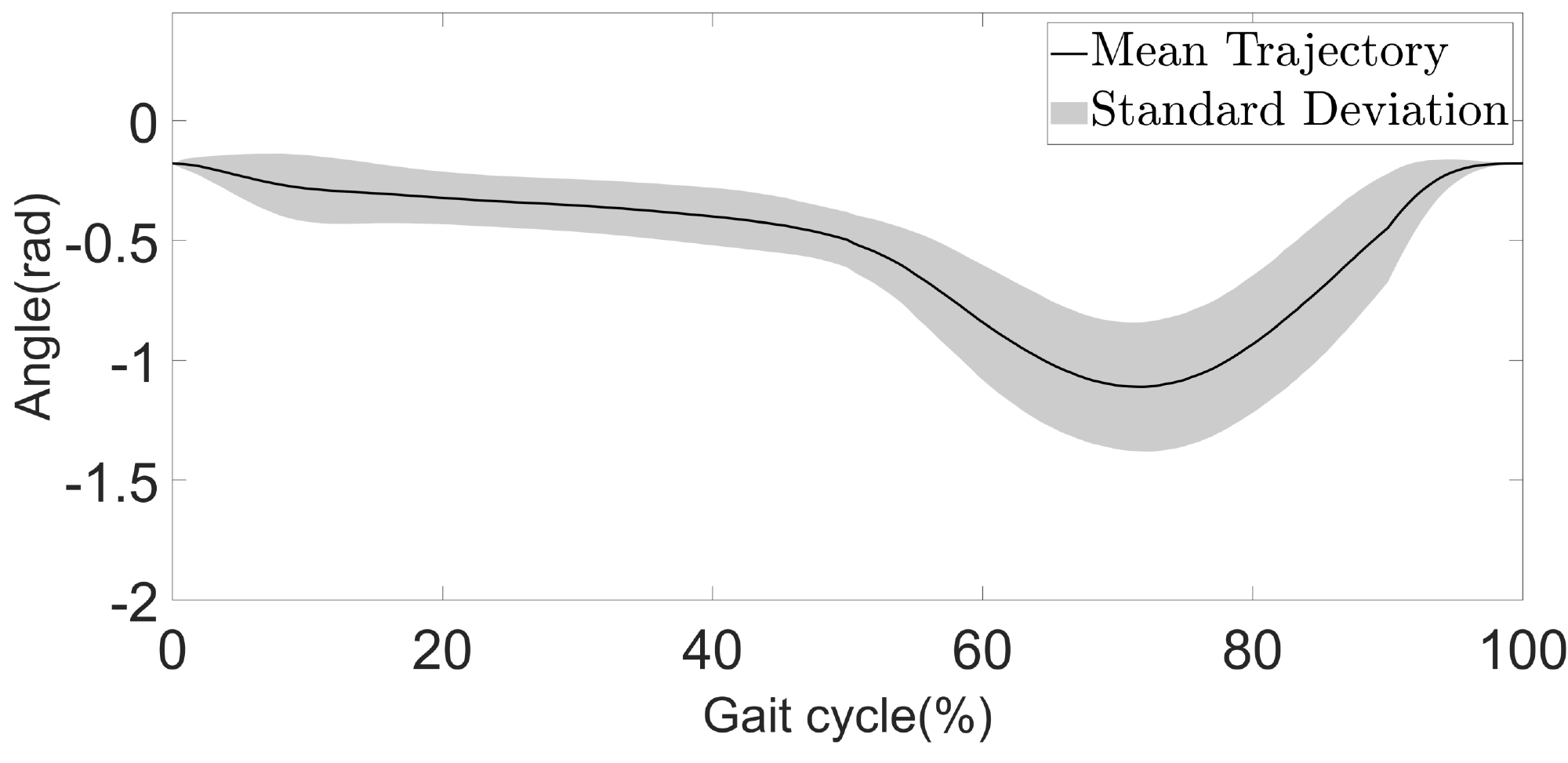}}
	\subfigure[]{
		\label{Up_curve}
		\includegraphics[width=0.32\linewidth]{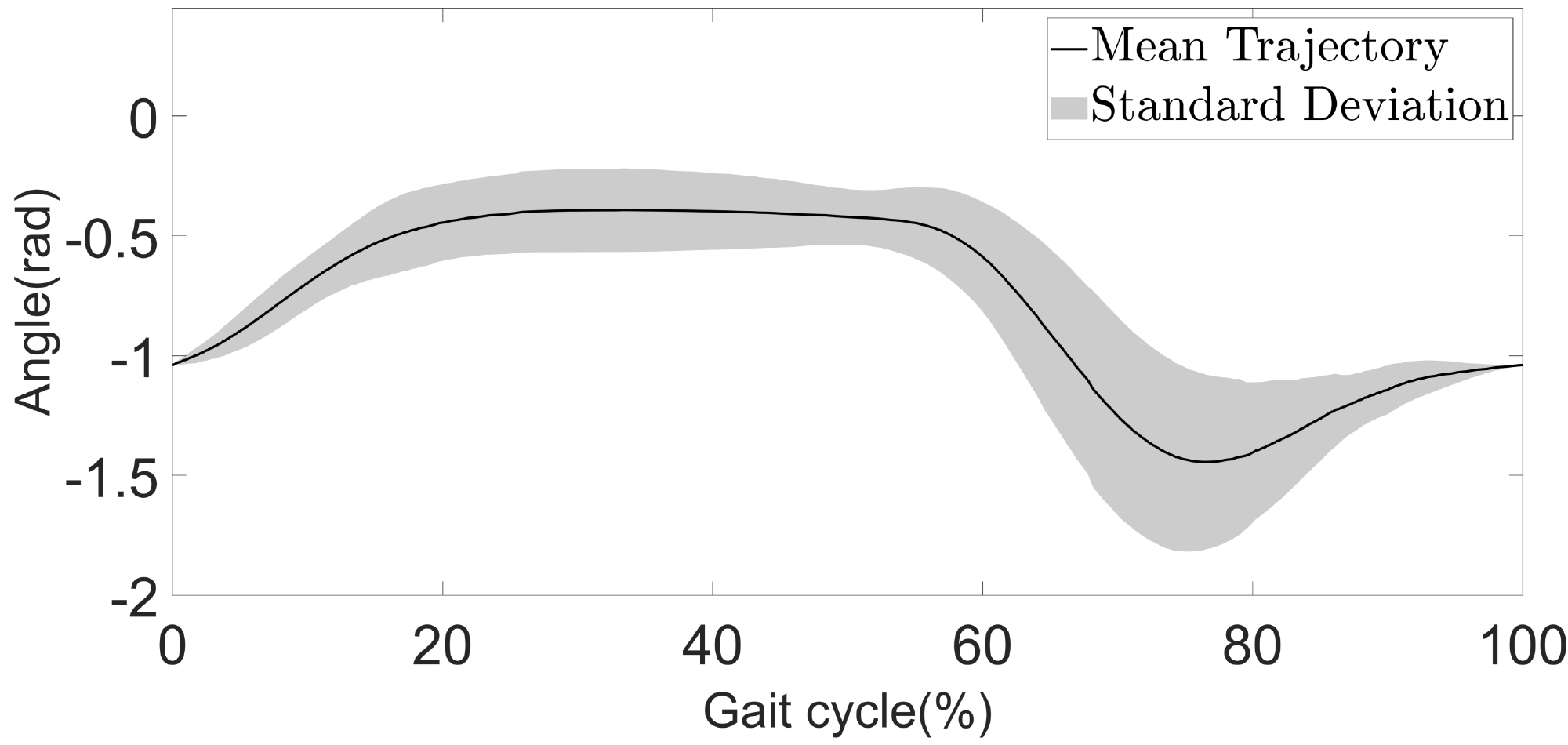}}
	\subfigure[]{
		\label{Down_curve}
		\includegraphics[width=0.32\linewidth]{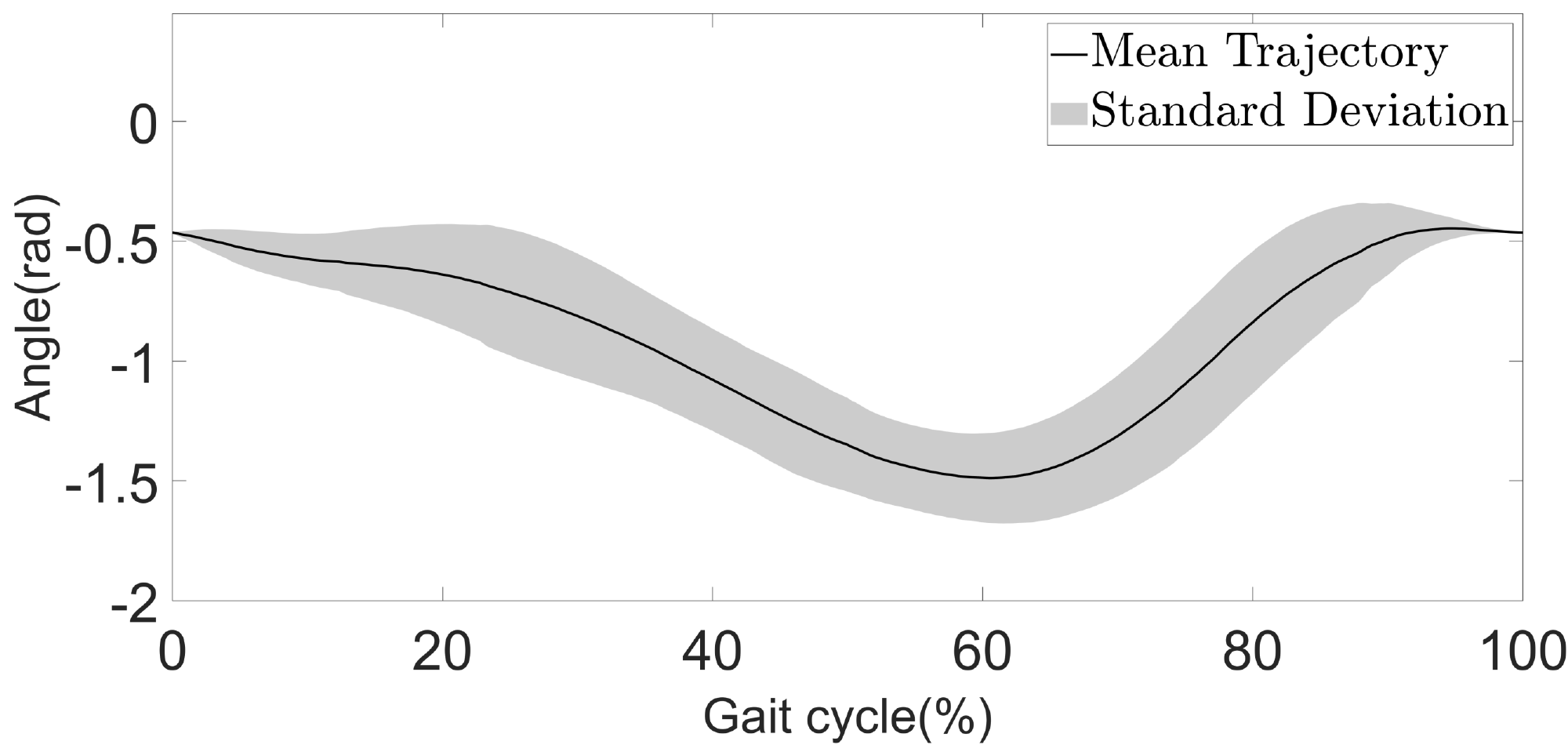}}\\
    \subfigure[]{
		\label{Sit_curve}
		\includegraphics[width=0.32\linewidth]{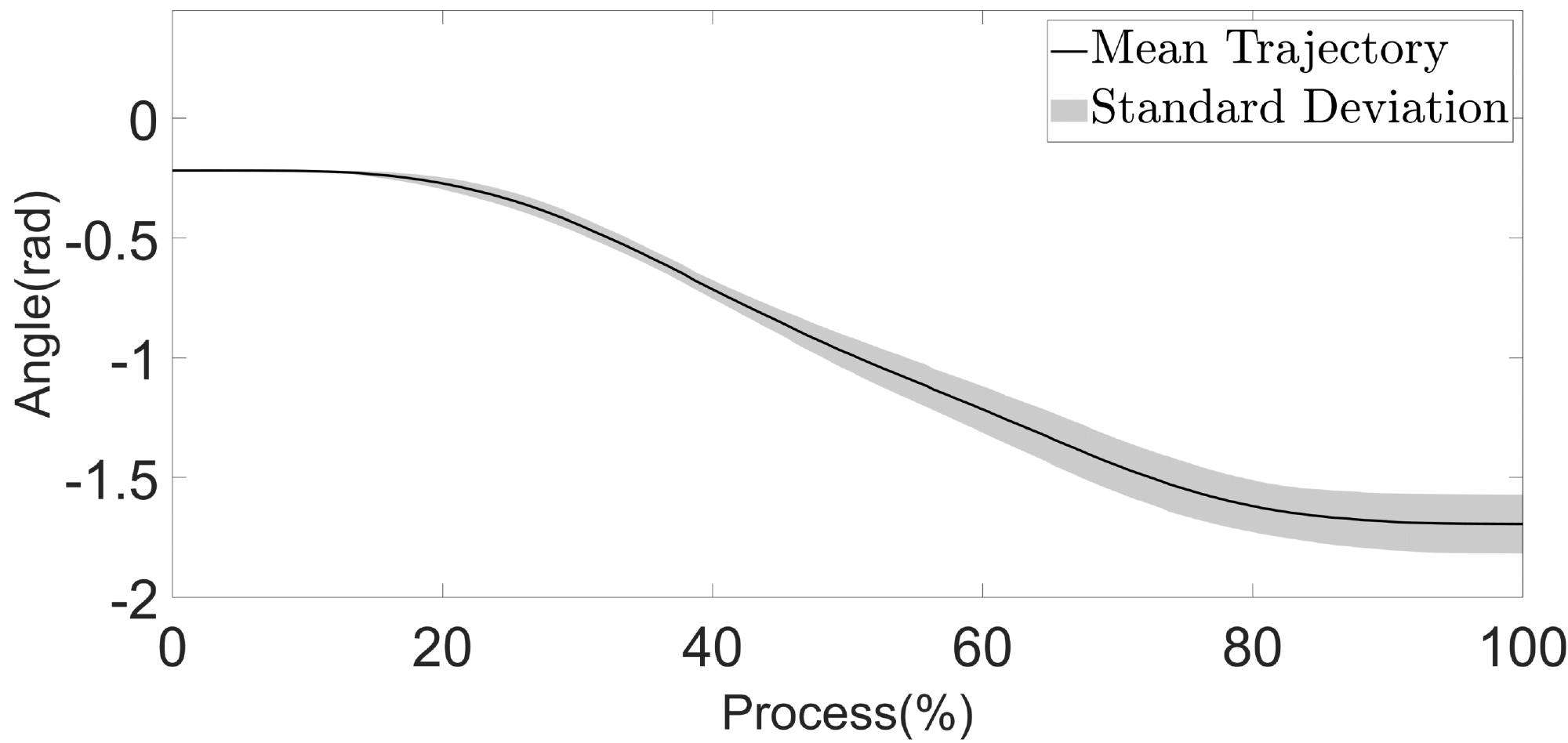}}
    \quad
    \subfigure[]{
		\label{Stand_curve}
		\includegraphics[width=0.32\linewidth]{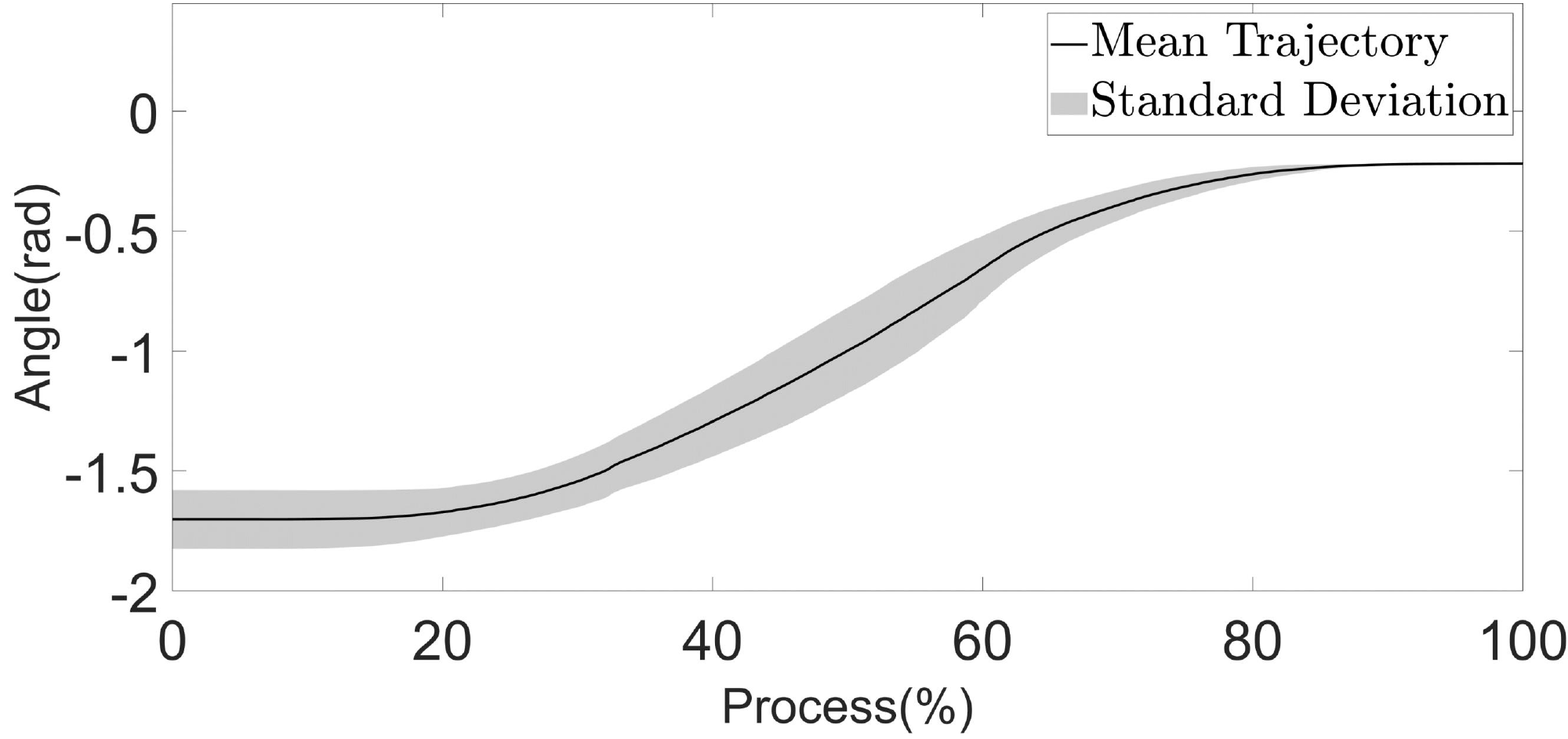}}
	\caption{Gait profiles of the left knee over different tasks: (a) Walking; (b) Ascending stairs; (c) Descending stairs; (d) Squatting; (e) Standing up. 
 The black solid line represents the mean of the dataset, with the surrounding black shade denoting its standard variation.}
	\vspace{0cm}
	\label{exp_curve_difftask}
\end{figure*}

An embedded main board facilitated communication with the PC, as shown in Fig. \ref{equipment}. 
The detection network was implemented on the PC, and the control algorithm was programmed and realized by the main board. The main board transmitted the control commands through the pulse width modulation (PWM) signals to the four slave boards to drive the motion of the four robot joints.
Furthermore, to minimize the influence of the motors on the wearer's movements during data collection and reduce variable dimensions in optimization, only the two motors located at the knee joint were activated in this experiment.

\subsection{Dataset of Human Subjects}
Ten able-bodied participants with no prior experience with exoskeletons performed multiple tasks while wearing the exoskeleton.
Table \ref{participants} presents an overview of the participants' statistical attributes, namely age ($25.6 \pm 1.9$ years), weight ($66.3 \pm 14.8$ kg), and height ($173.4 \pm 9.1$ cm).
The experimental protocol was approved by the ethics committee initiated by Shenzhen MileBot Robotics Co., Ltd in May 2023.
All participants signed a written informed consent form prior to the experimental sessions.

\begin{table}[h]
\caption{Statistical information of the subjects}
\centering
\begin{tabular}{ccccc} 
\toprule
Subject & Gender  & Age(y) & Weight(kg) & Height(cm)  \\ 
\midrule
1   & Male   & 25   & 81        & 185        \\
2 & Male   & 25   & 70        & 178        \\
3 & Male   & 26   & 78        & 177        \\
4 & Male   & 24   & 57        & 172        \\
5 & Male   & 25   & 74        & 183        \\
6 & Male   & 30   & 87        & 181        \\
7 & Male   & 27   & 72        & 172        \\
8 & Female   & 23   & 46        & 159        \\
9 & Female   & 26   & 48        & 160        \\
10 & Female   & 25   & 50        & 167        \\
\bottomrule
\end{tabular}
\label{participants}
\end{table}



As shown in Fig. \ref{snapshot_task}, the participants were instructed to perform five tasks at their own pace: walking, ascending stairs, descending stairs, squatting, and standing up, as summarized in Table \ref{tasks_intro}.
Throughout data collection, the exoskeleton operated in a transparent mode to allow the wearer to comfortably perform multiple tasks, and the motion frequency was recorded using (\ref{eq:omega}). 
During squatting and standing tasks, the exoskeleton applied a constant torque to support the wearer, with the magnitude of the torque being meticulously adjusted for each individual to ensure comfortable movement. Additionally, each participant underwent a six-minute pre-training session to become familiar with
each task. 
Fig. \ref{exp_curve_difftask} shows the angle profiles of the left knee joint under different tasks. 
The constructed dataset was used to train the anomaly detection and task translation modules.

\begin{table}[h]
\caption{Tasks}
\centering
\begin{tabular}{l|l}
\hline
\textbf{Task} & \textbf{Description} \\ \hline
Walking & To walk at a comfortable pace. \\ \hline
Ascending stairs & To ascend a flight of stairs at a comfortable pace. \\ \hline
Descending stairs & To descend a flight of stairs at a comfortable pace. \\ \hline
Squatting & To squat and hold the position for a brief period. \\ \hline
Standing up & To stand up from a seated position. \\ \hline
\end{tabular}
\label{tasks_intro}
\end{table}


\subsection{Anomaly Detector}
Before training the anomaly detector, the min-max normalization technique was applied to reformat the sensory data of the exoskeleton robot and ensure scale uniformity. 
When the participant wearing the exoskeleton could easily perform the task, the collected data were considered normal.
Otherwise, they were labeled as abnormal due to physical conflicts between the human and robot.
\begin{figure}[!h] 
	\centering 
	\vspace{-0.3cm} 
	\subfigtopskip=2pt 
	\subfigbottomskip=2pt 
	\subfigcapskip=-5pt 
	\subfigure[]{
		\includegraphics[width=0.31\linewidth]{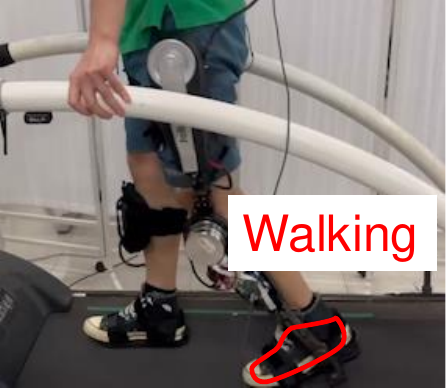}}
	\subfigure[]{
		\includegraphics[width=0.31\linewidth]
        {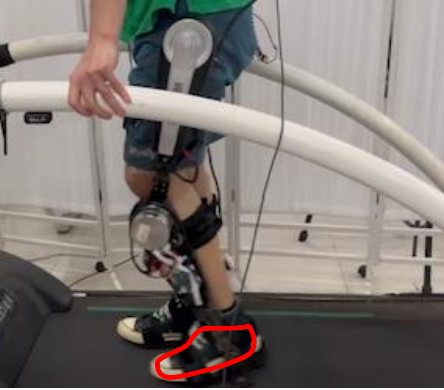}}
	\subfigure[]{
		\includegraphics[width=0.31\linewidth]{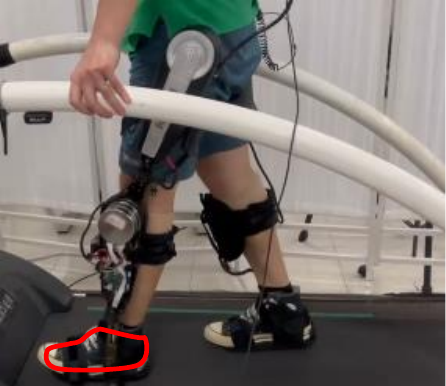}}
    \subfigure[]{
		\includegraphics[width=0.31\linewidth]{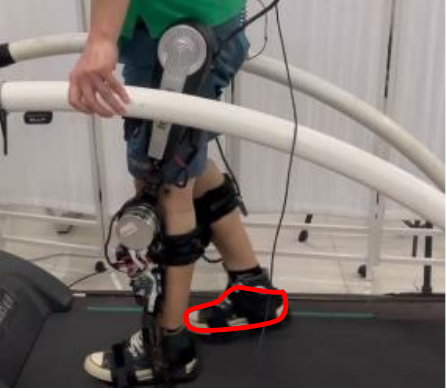}}
    \subfigure[]{
		\includegraphics[width=0.31\linewidth]{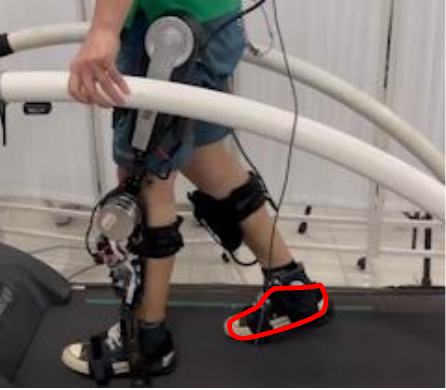}}
   \subfigure[]{
		\includegraphics[width=0.31\linewidth]{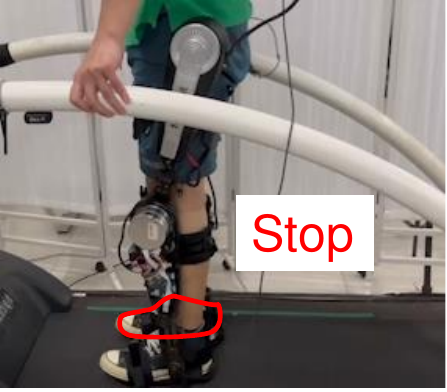}}
  \subfigure[]{
		\includegraphics[width=1\linewidth]
  {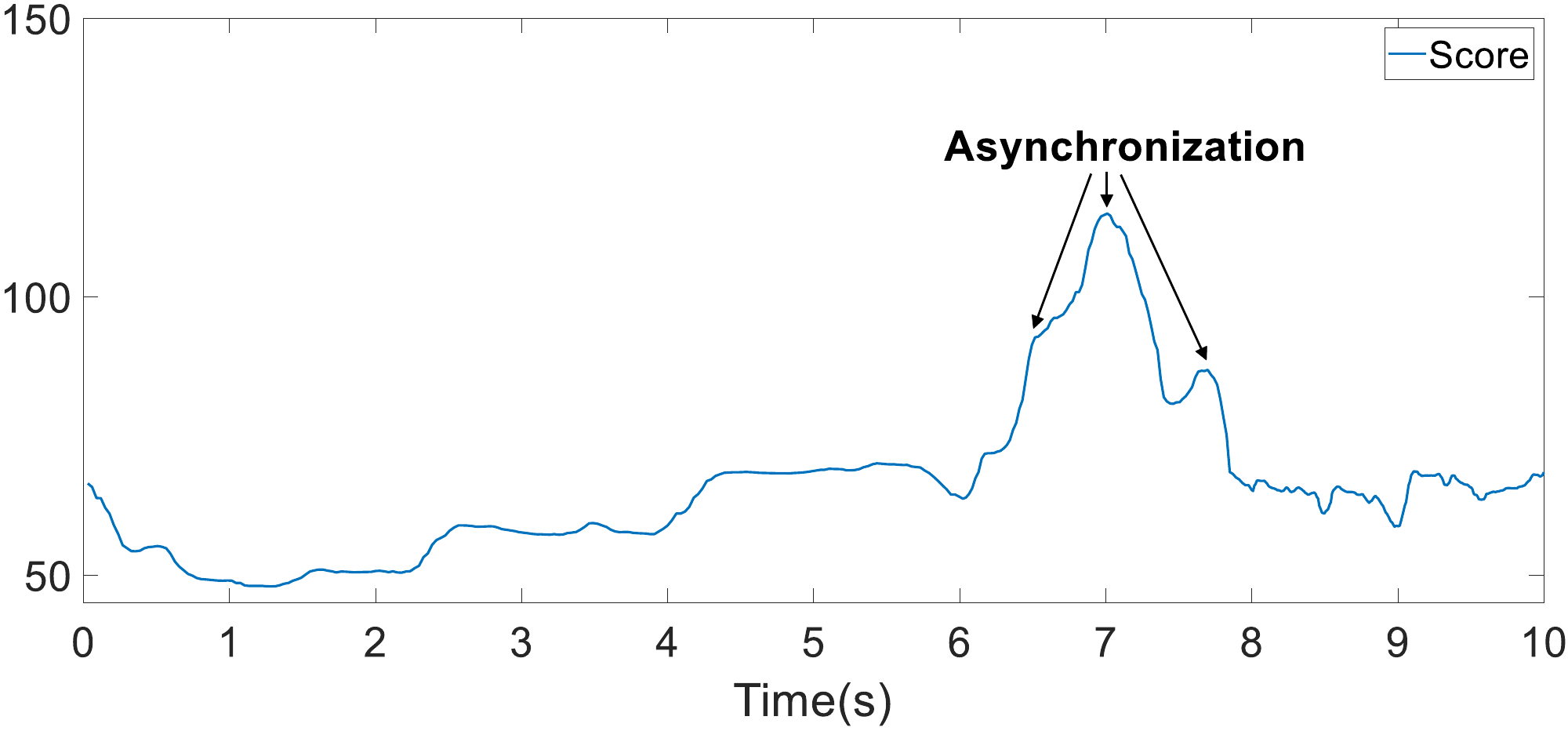}}
	\caption{(a)-(f) Snapshots of conflict due to asynchronization; (g) anomaly score during asynchronization. The red contour represents the foot during the stepping motion.}
	\label{asy_ano}
\end{figure}

\begin{figure}[!h] 
	\centering 
	\vspace{-0.3cm} 
	\subfigtopskip=2pt 
	\subfigbottomskip=2pt 
	\subfigcapskip=-5pt 
	\subfigure[]{
		\includegraphics[width=0.31\linewidth]{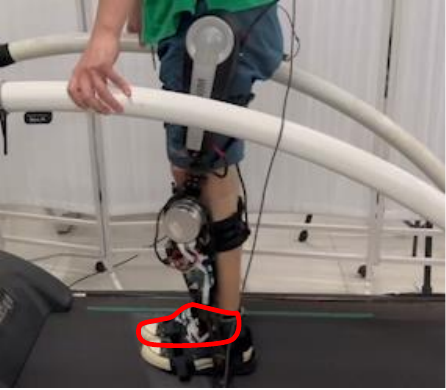}}
	\subfigure[]{
		\includegraphics[width=0.31\linewidth]
        {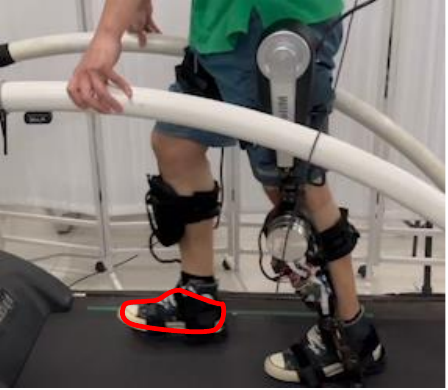}}
	\subfigure[]{
		\includegraphics[width=0.31\linewidth]{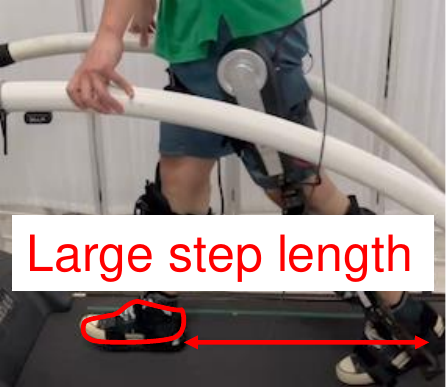}}
    \subfigure[]{
		\includegraphics[width=0.31\linewidth]{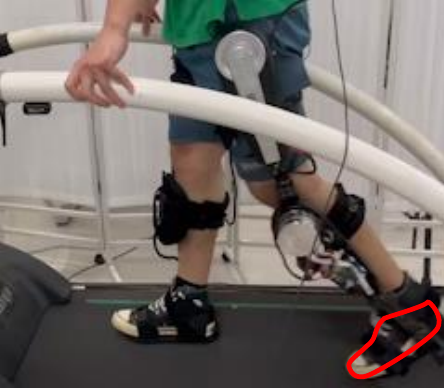}}
    \subfigure[]{
		\includegraphics[width=0.31\linewidth]{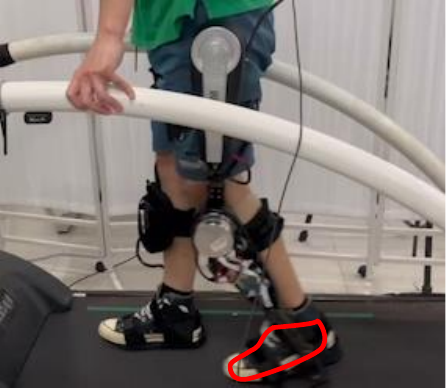}}
   \subfigure[]{
		\includegraphics[width=0.31\linewidth]{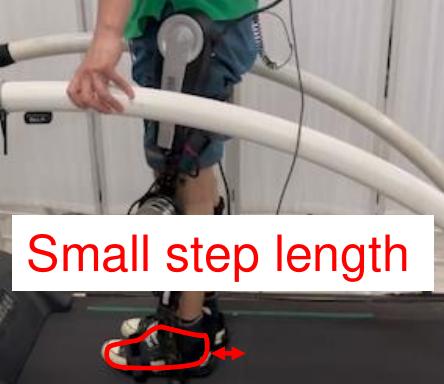}}
  \subfigure[]{
		\includegraphics[width=1\linewidth]
  {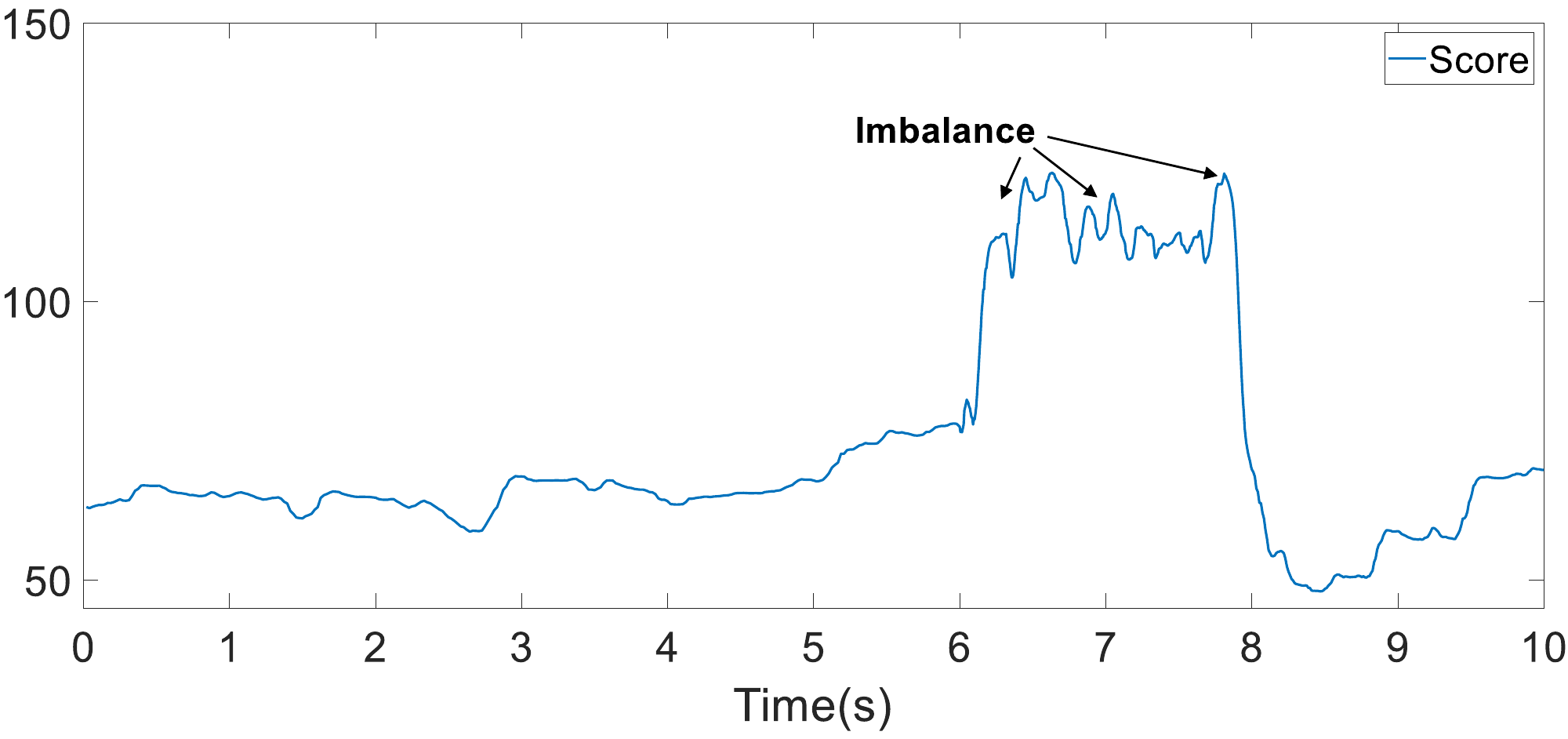}}
	\caption{(a)-(f) Snapshots of conflict due to imbalance; (g) anomaly score during imbalance. The red contour represents the foot during the stepping motion.}
	\label{imb_ano}
\end{figure}

During the experiment, a participant intentionally held the robot to simulate conflict.
The output of the detection network in this condition is shown in Fig. \ref{asy_ano}. 
The reconstruction error 
significantly increased when the conflict arose, indicating the wearer's discomfort.
This finding highlighted that the anomaly could be successfully detected.
Furthermore, the performance of the detection network was validated in another simulated conflict, where the participant lost balance due to fatigue. The results are shown in Fig. \ref{imb_ano}, further reinforcing the network's capacity to successfully detect such conflicts.

\begin{figure}[!h]
    \vspace{-0.5cm}
    \centering
      \subfigure[ ]{
          \label{roc1}
          \includegraphics[width=4.0cm]{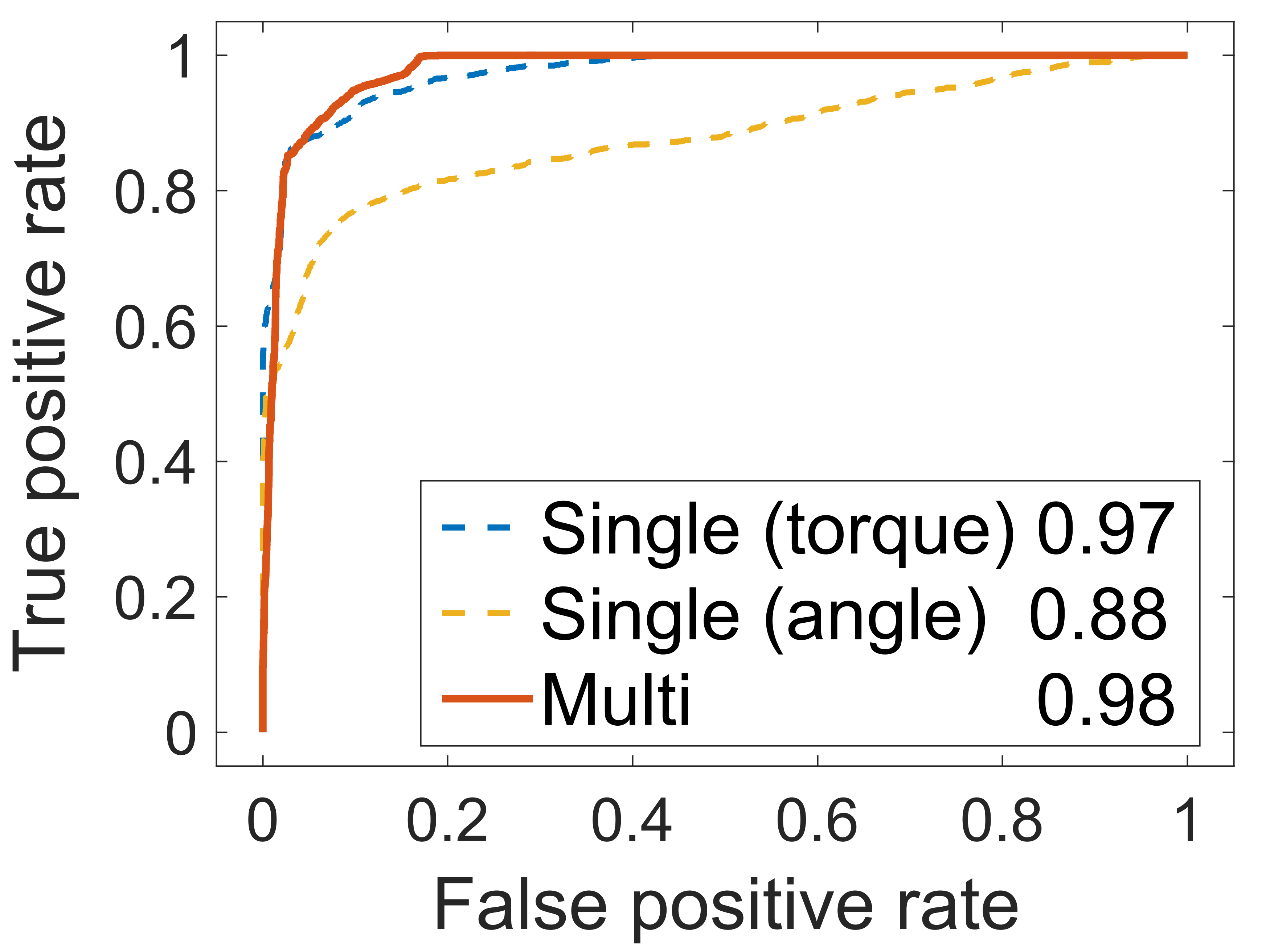}
          }\hfill
      \subfigure[ ]{
          \label{roc2}
          \includegraphics[width=4.0cm]{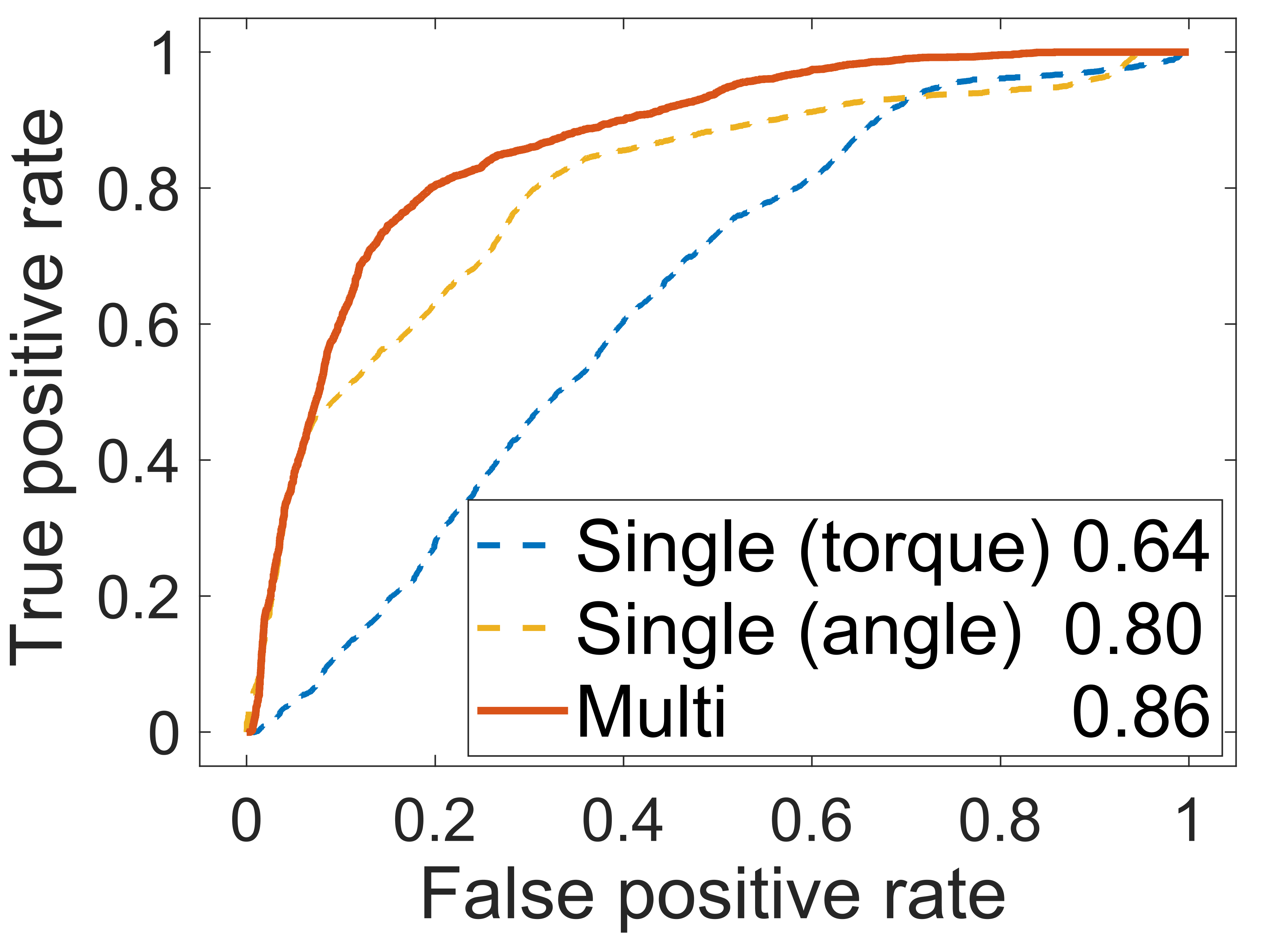}
          } 
\caption{AUC values (shown in the legend) and the ROC curve of multi-modal detection in comparison with single input systems (using only the torque or encoder information) in two scenarios: (a) asynchronization; (b) imbalance.}
\label{ROC_curve}
\vspace{-0.15cm}
\end{figure}

In addition, ablation studies were performed to show that the use of the force information provided by the SEA structure could enhance the detection accuracy compared with that associated with single sensory input.
The comparison results are shown in Fig. \ref{ROC_curve}. 
The detection performance was evaluated using the receiver operating characteristic (ROC) curve, which indicates how well a classification model can distinguish classes.
In general, a larger area under the curve (AUC) indicates better performance.
Fig. \ref{ROC_curve} shows that the anomaly detection network incorporating both angular and force information demonstrated a higher AUC than systems relying on single sensory input across two scenarios(i.e., asynchronization and imbalance). 
This finding highlights that the proposed anomaly detection network enables more precise detection, owing to its ability to handle diverse anomaly scenarios with specific manifestations.
For instance, in the presence of asynchronization between the human and robot, the interaction torque rapidly increased. 
Conversely, in cases of imbalance due to human fatigue, angular information on the human limb, such as encoder signals, changed more significantly.

\subsection{Task Translator}
To accommodate trajectories across different tasks, the collected data were used to train a task translator. This process was modeled as a regression problem to promote generalization across diverse tasks. Owing to the limited sample size, we used a leave-one-out cross-validation approach and applied various techniques to address the regression problem, namely ridge regression, KNN \cite{peterson2009k}, GPR \cite{williams1995gaussian}, XGBoost \cite{chen2016xgboost}, GBoost \cite{friedman2001greedy}, SVM \cite{hearst1998support}, and NN.

\begin{figure*}[!ht] 
	\centering 
	\vspace{-0.3cm} 
	\subfigtopskip=2pt 
	\subfigbottomskip=2pt 
	\subfigcapskip=-5pt 
	\subfigure[]{
		\label{Up_traj}
		\includegraphics[width=0.45\linewidth]{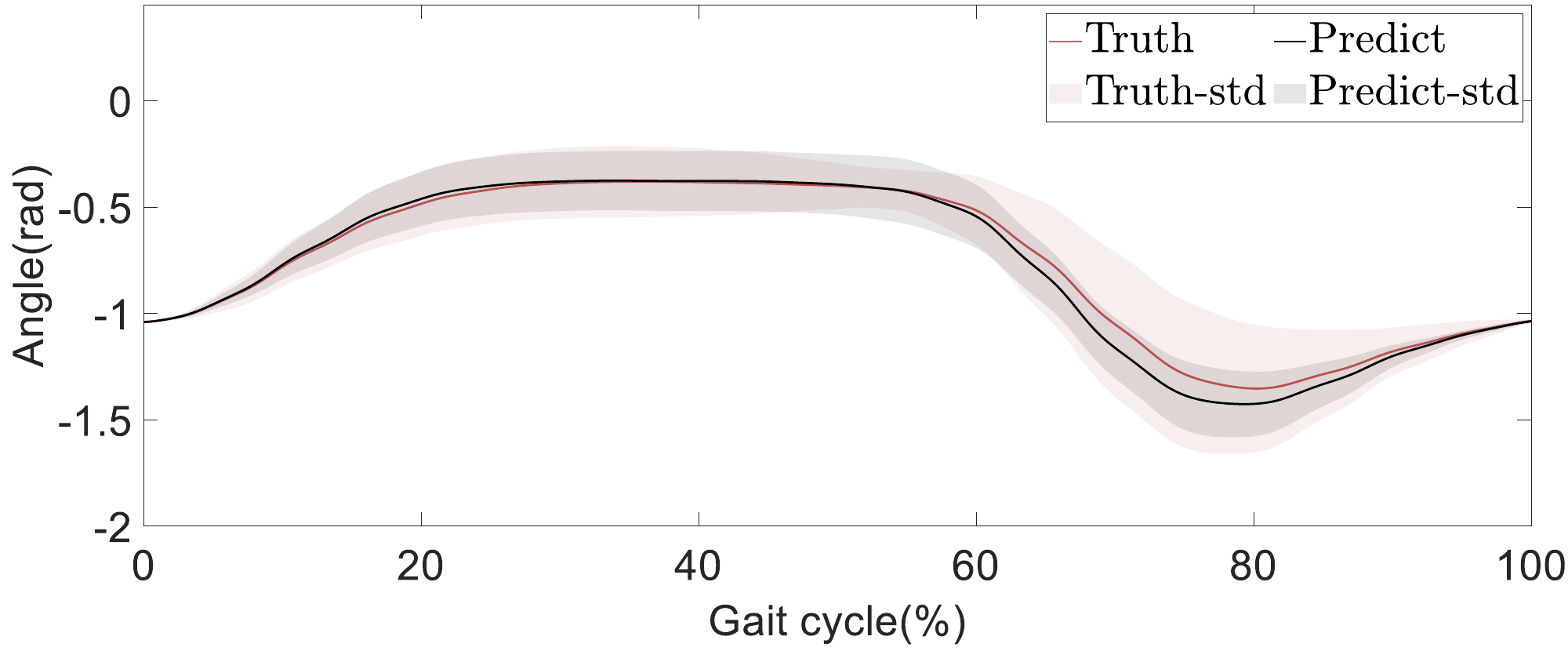}}
	\subfigure[]{
		\label{Down_traj}
		\includegraphics[width=0.45\linewidth]{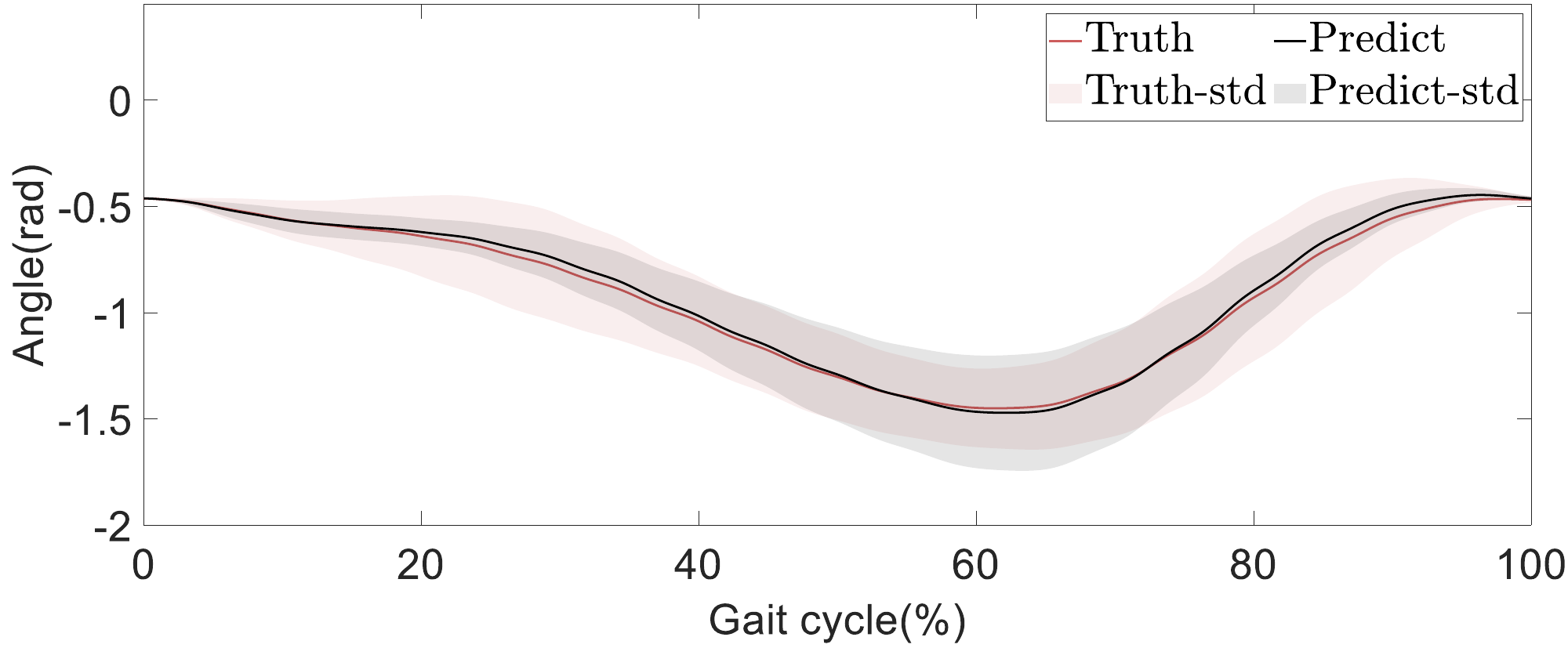}}\\
    \subfigure[]{
		\label{Sit_traj}
		\includegraphics[width=0.45\linewidth]{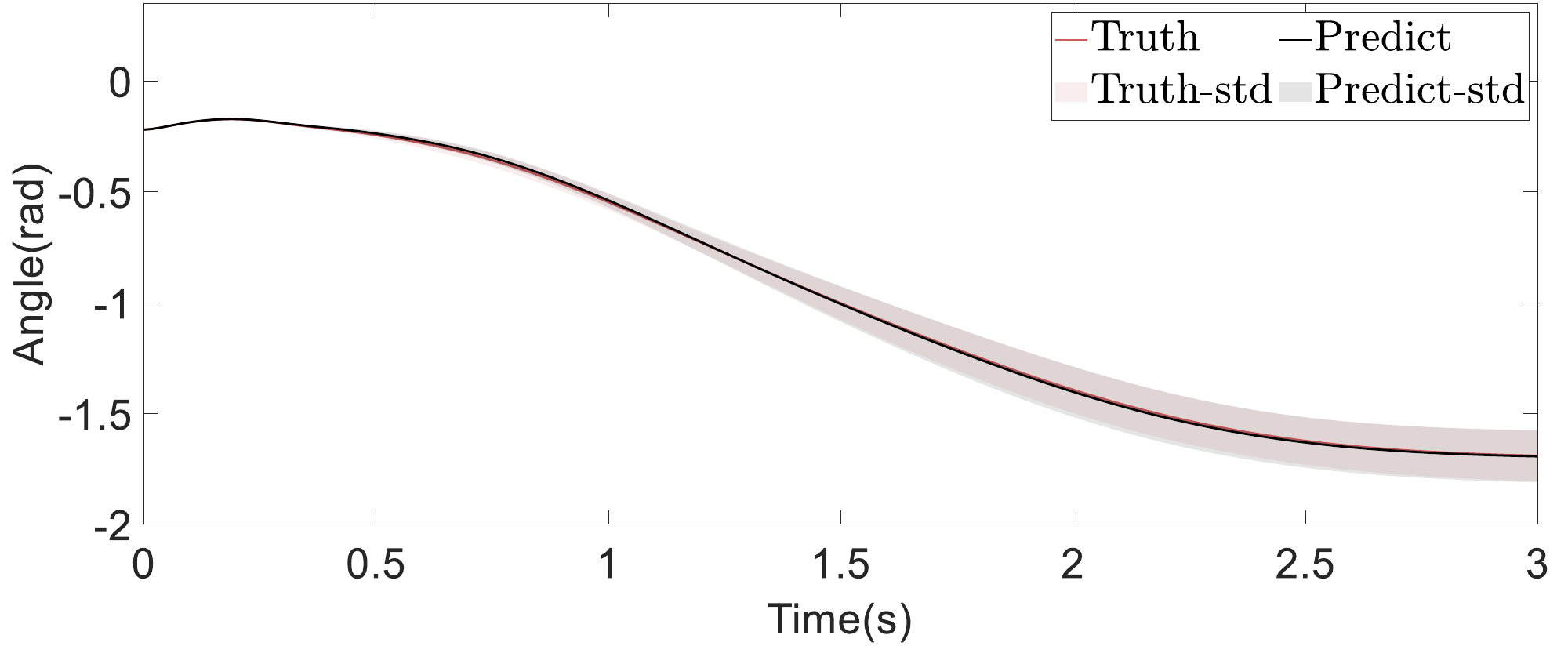}}
    \quad
    \subfigure[]{
		\label{Stand_traj}
		\includegraphics[width=0.45\linewidth]{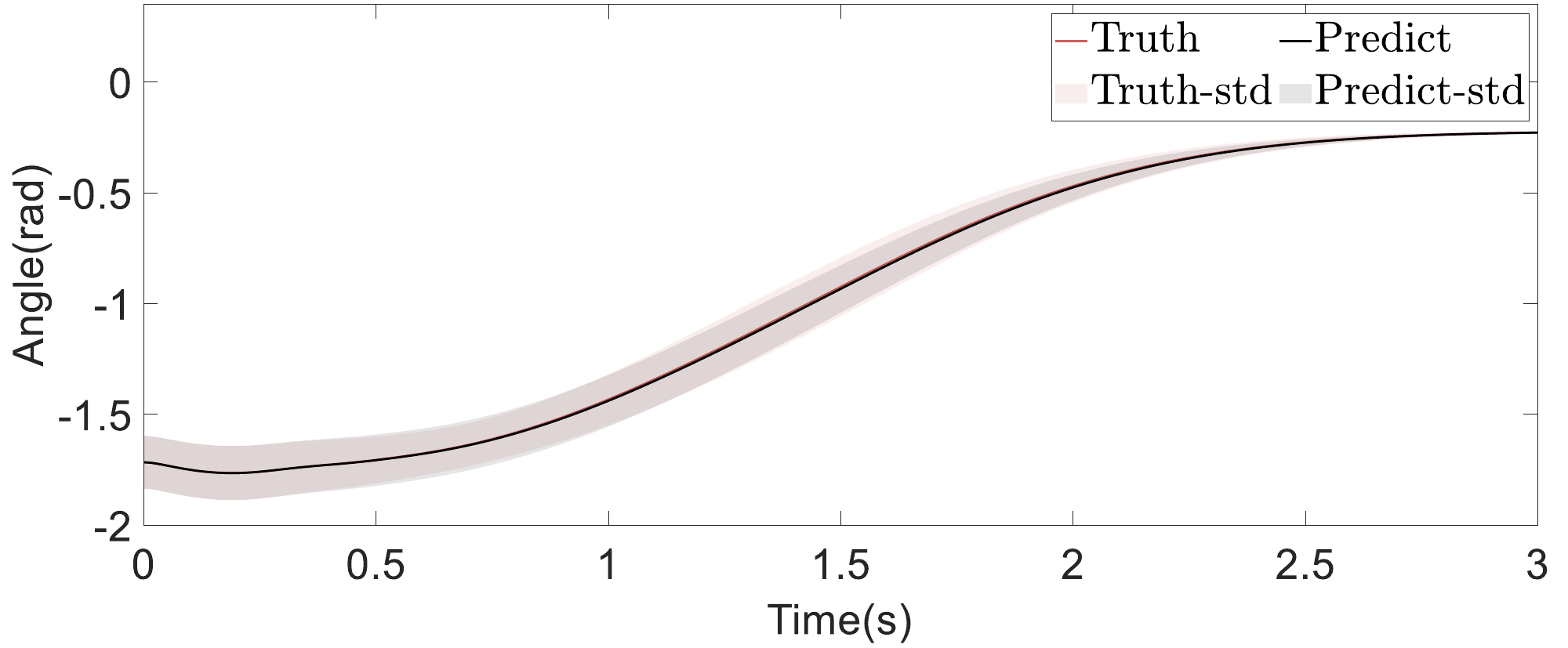}}
	\caption{Translation results across different tasks: (a) Ascending stairs; (b) Descending stairs; (c) Squatting; (d) Standing up.
 The red solid line represents the mean of the true profile, with the red shade indicating its standard variation. Conversely, the black solid line indicates the prediction from the task translator, with the surrounding black shade denoting its standard variation.}
	\vspace{0cm}
	\label{trans_traj}
\end{figure*}

Trajectories were encoded into weight matrices corresponding to distinct tasks to allow these techniques to predict the transformation of the weight matrix from one task to another, for example, the transition from walking to ascending stairs. 
The root mean square error (RMSE) within the joint space was used as the evaluation metric for this regression task.
The RMSE values are expressed in degree units.
A comparison of the translation from walking to ascending stairs, using the aforementioned techniques, is presented in Table \ref{result_trans}.

Note that the complexity of the regression task increases in tandem with an increase in the number of kernels. 
In addition, among the compared models, the NN-based translator demonstrated superior performance. Considering both the fitting error and expression of DMP, the task translator included the NN, and the optimal number of kernels was determined to be $J=20$.


\begin{table}
\caption{translation results for ascending stairs}
\centering
\begin{tabular}{cccccccc} 
\toprule
$J$ & Ridge & KNN & GPR & XGBoost & GBoost & SVM & NN \\ 
\midrule
15   & 4.74   & 5.12  & 5.34 & 6.57   & 5.61   & 4.89   &  \bf{2.94}  \\
20   & 4.81   & 5.19  & 5.31 & 6.44   & 5.63   & 5.06   &  \bf{3.33}  \\
25   & 4.87   & 5.23  & 5.39 & 6.74   & 5.91   & 5.08   &  \bf{3.97}  \\
\bottomrule
\multicolumn{8}{l}{\textit{Note:} The RMSE values are presented in degree units.}
\end{tabular}
\begin{tablenotes}
\small
\item[1] $J$ is the kernel number for rhythmic movement.
\end{tablenotes}
\label{result_trans}
\vspace{-0.3cm}
\end{table}

\begin{table}
\caption{translation results for standing}
\centering
\begin{tabular}{cccccccc} 
\toprule
$J_d$ & Ridge & KNN & GPR & XGBoost & GBoost & SVM & NN \\ 
\midrule
10   & 4.54   & 4.60 & 4.69   & 4.35   & 4.55   & 4.43   &  \bf{2.71}  \\
15   & 4.81   & 4.87 & 4.86   & 4.66   & 4.86   & 4.69   &  \bf{2.89}  \\
20   & 4.95   & 5.01 & 4.97 & 4.83   & 4.75   & 4.87   &  \bf{2.95}  \\
\bottomrule
\multicolumn{8}{l}{\textit{Note:} The RMSE values are presented in degree units.}
\end{tabular}
\begin{tablenotes} 
\small
\item[1] $J_d$ is the kernel number for discrete movement.
\item[2] The number of kernels for rhythmic movement is set as  $J=20$.
\end{tablenotes}
\label{result_trans_discrete}
\vspace{-0.5cm}
\end{table}

Upon ascertaining the kernel number for the rhythmic motion, the same approaches were used to execute the translation into discrete movements, specifically, the transitions between the squatting and standing postures. The translation outcomes for standing are summarized in Table \ref{result_trans_discrete}. 
Among the evaluated methods, the NN outperformed the other techniques and was thus selected for the translation of these discrete movements. Considering both the fitting error and DMP expression, the final kernel number was selected as $J_d = 15$.


Hence, within the framework of the task translator, we selected the NN for translation. The translation results, derived from the application of the leave-one-out cross-validation, are visualized in Fig.\ref{trans_traj}.
Note that the reproduced profile aligns satisfactorily with the collected gait profile across different tasks, despite the limitations of a small dataset.
This is evident in the metric of the average RMSE, which yields values of $3.33$ for ascending stairs, $2.64$ for descending stairs, $2.93$ for squatting, and $2.89$ for standing up.

\begin{figure}[!h]
    \centering
    \includegraphics[width=0.6\linewidth]{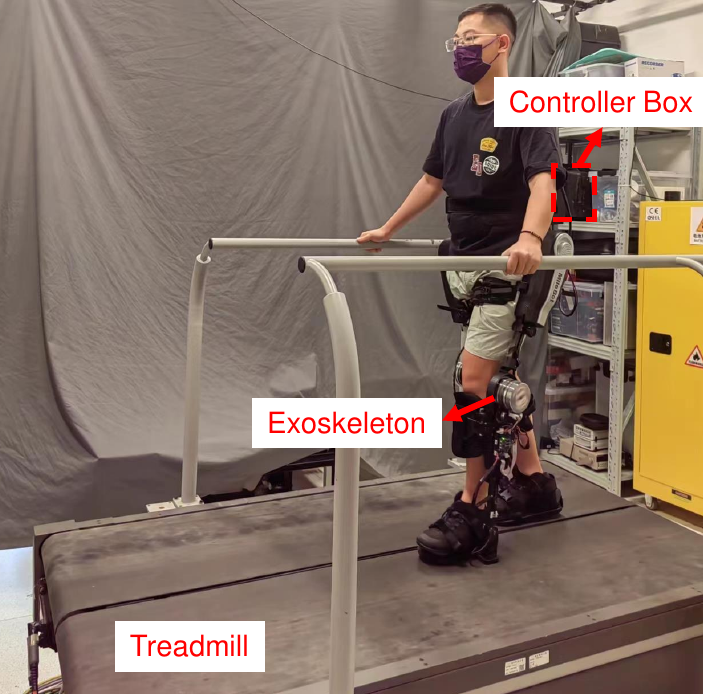}
    \caption{Experimental setup for HIL optimization, where the participant walks on a treadmill with the exoskeleton.}
    \label{snap_HIL}
\end{figure}

\subsection{HIL Optimization}
\begin{figure*}[!ht]
    \centering
    \includegraphics[width=14cm]{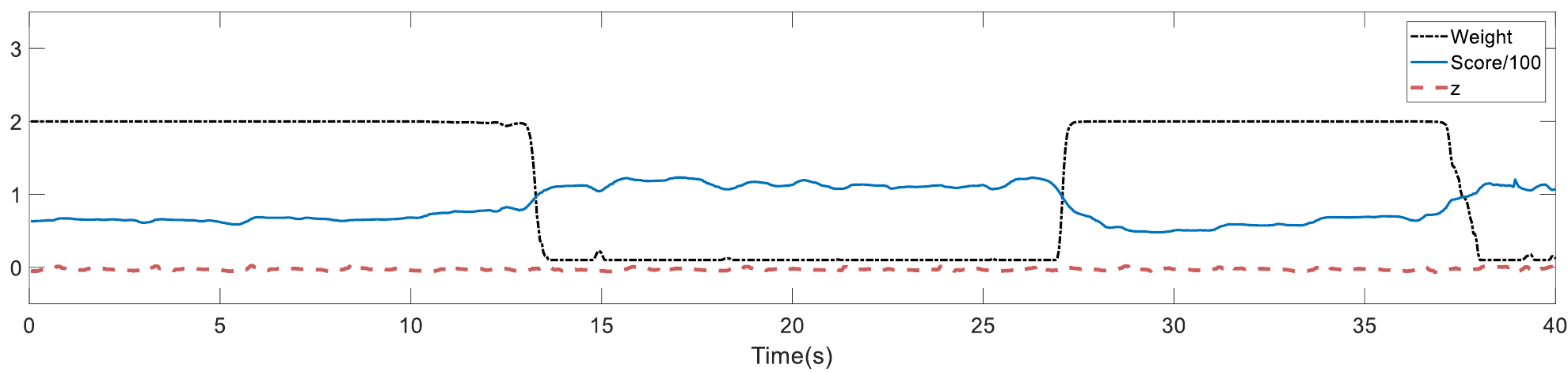}
    \caption{Experimental results using the proposed variable impedance controller: The impedance parameters were scaled down (black dashed line) when the anomaly score increased (blue solid line), thereby alleviating the conflicts. The adjustment of impedance is continuous as the change in $w(s)$ is smooth, ensuring safe human-robot interaction. The impedance vector (red dashed line) remained zero, proving the realization of the variable impedance model.}
    \label{Variable Impedance Weight}
\end{figure*}

\begin{figure*}[!ht] 
	\centering 
	\vspace{-0.3cm} 
	\subfigtopskip=2pt 
	\subfigbottomskip=2pt 
	\subfigcapskip=-5pt 
	\subfigure[]{
		\label{zx_HIL}
		\includegraphics[width=0.45\linewidth]{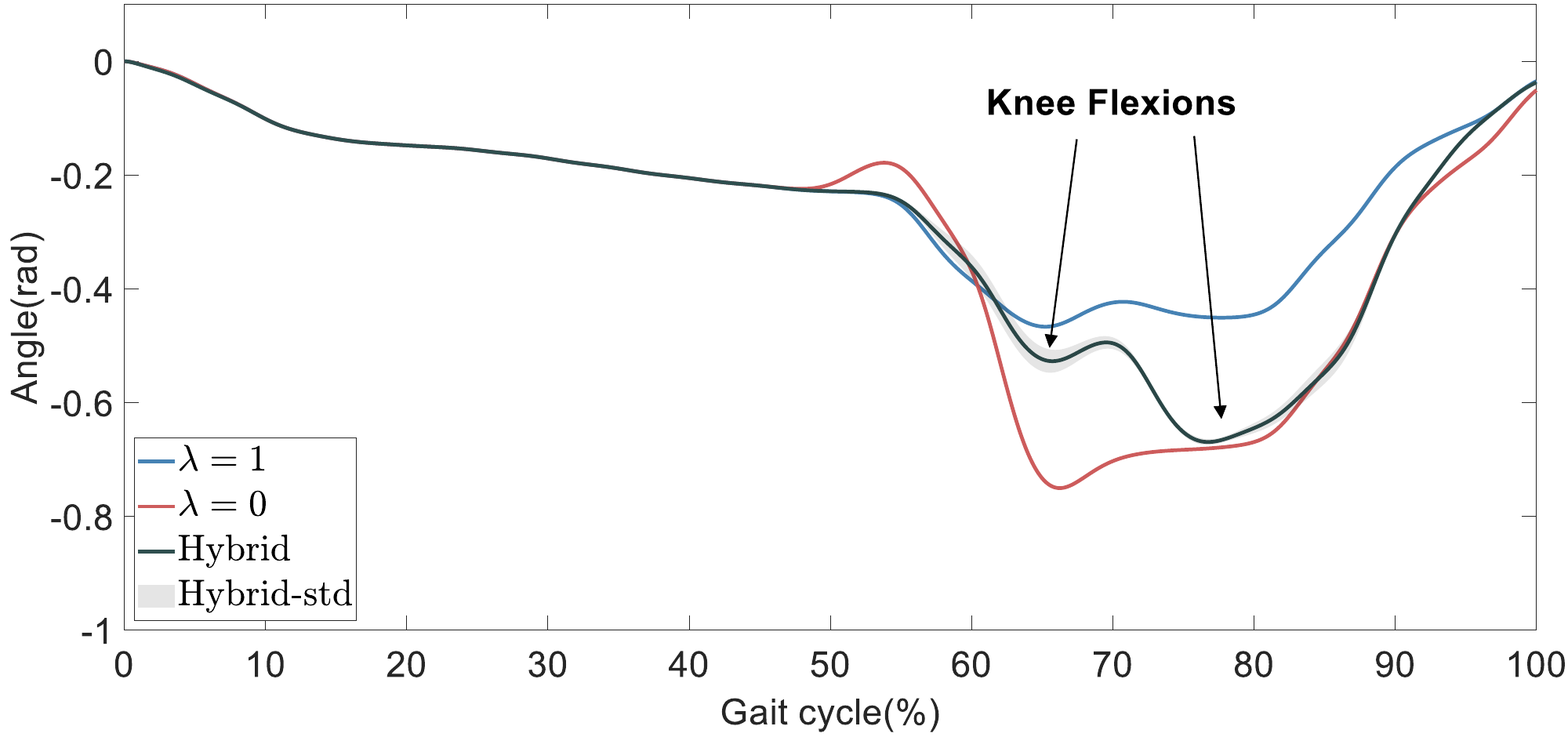}}
	\quad 
	\subfigure[]{
		\label{ms_HIL}
		\includegraphics[width=0.45\linewidth]{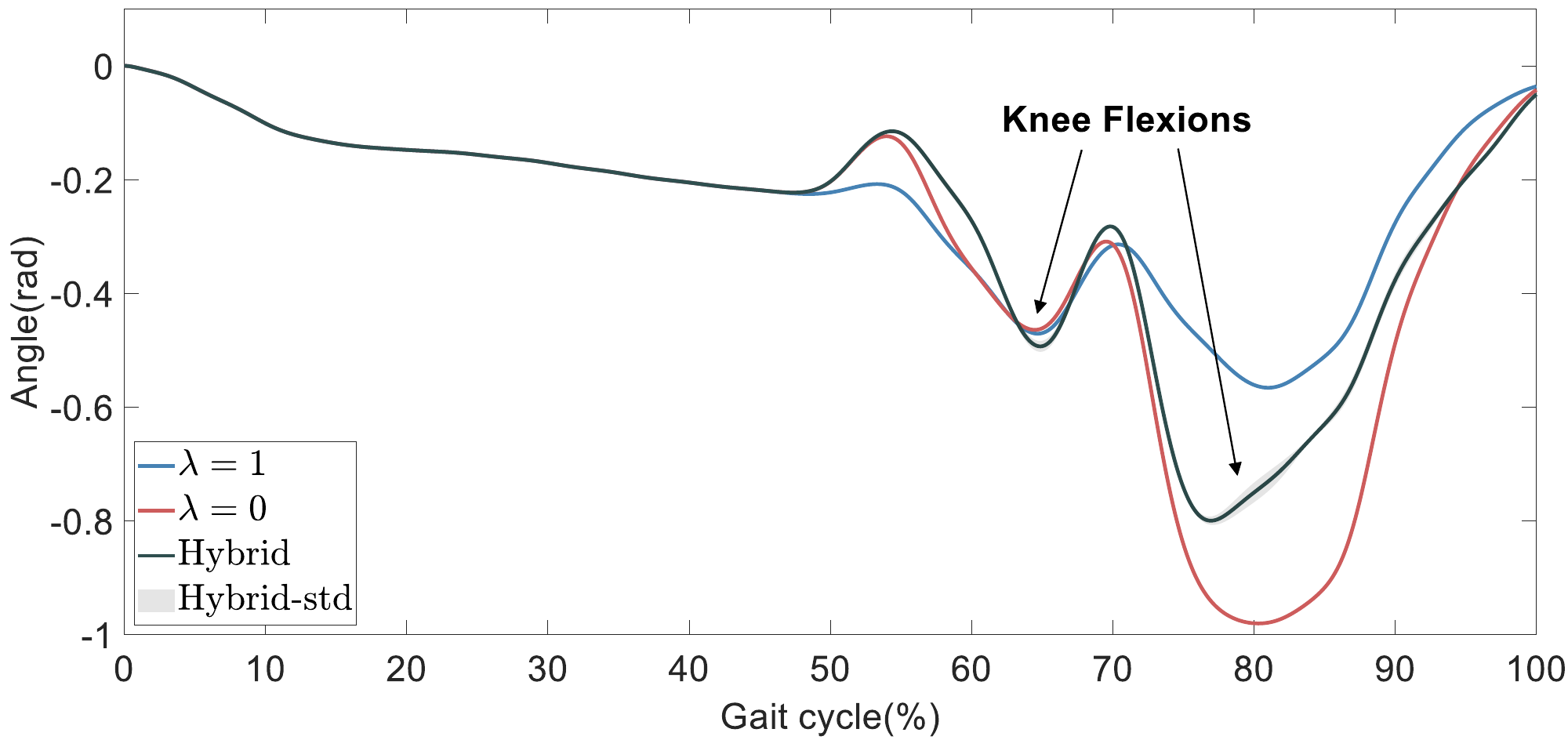}}
	\caption{HIL optimization results of different hyperparameters $\lambda$ for different participants: (a) Participant 1; (b) Participant 2.
 The blue and red solid lines represent the optimized trajectories in the context of $\lambda = 1$ and $\lambda = 0$, respectively. The black solid line is the mean of the trajectories, with $\lambda \in (0,1)$, and the surrounding black shade indicates its standard variation. }
	\vspace{0cm}
	\label{HIL_graph}
\end{figure*}

\begin{figure*}[!ht] 
	\centering 
	\vspace{-0.3cm} 
	\subfigtopskip=2pt 
	\subfigbottomskip=2pt 
	\subfigcapskip=-5pt 
	\subfigure[]{
		\label{walkingTA}
		\includegraphics[width=0.32\linewidth]{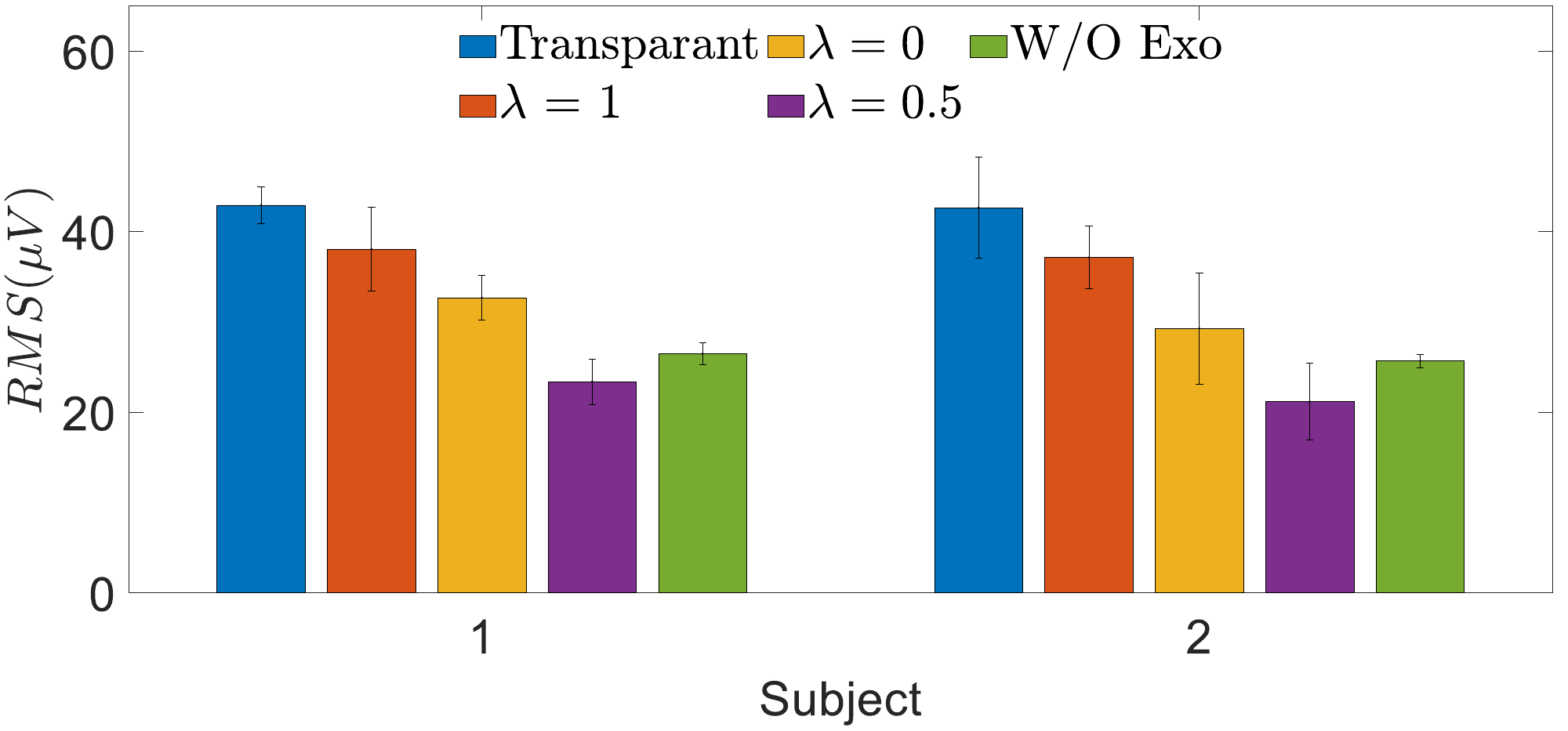}}
	\subfigure[]{
		\label{walkingQF}
		\includegraphics[width=0.32\linewidth]{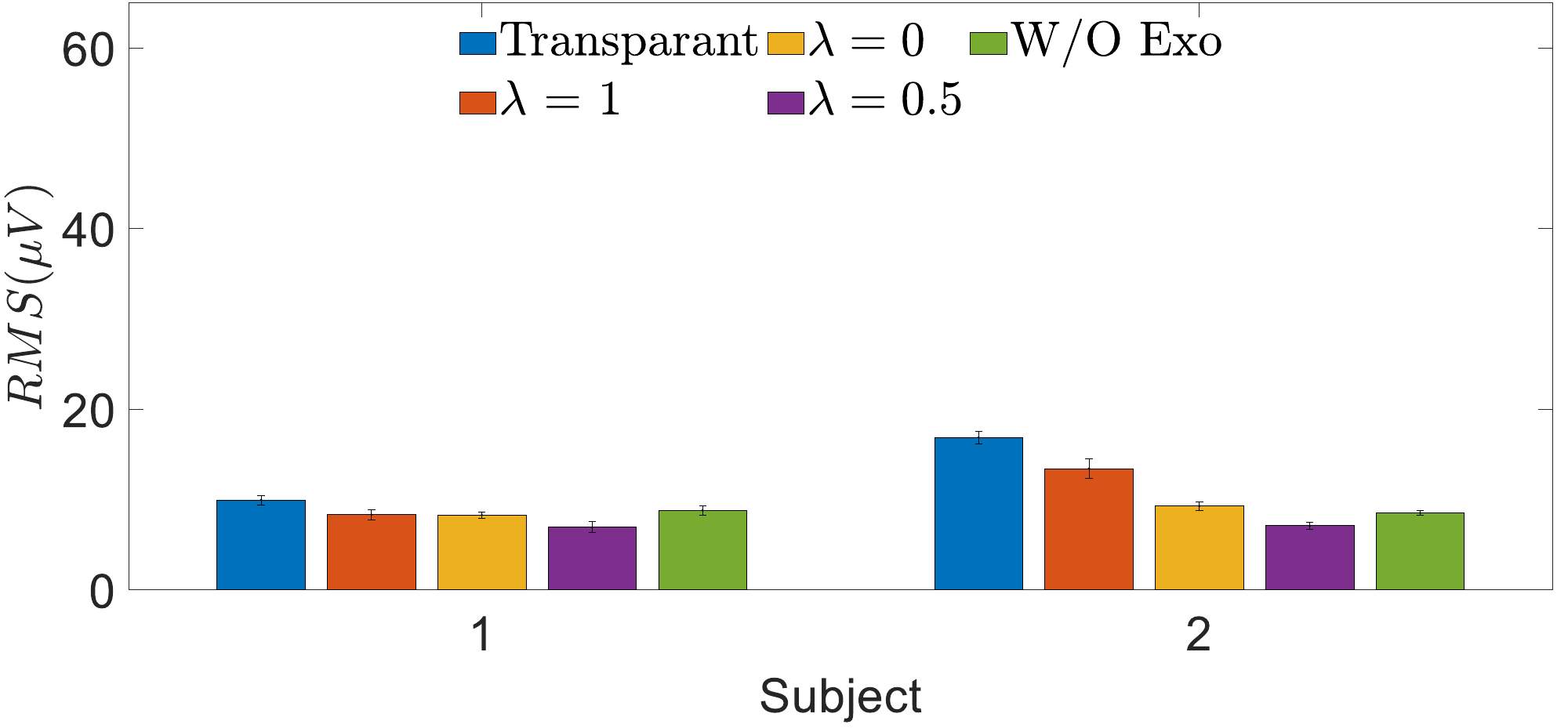}}
	\subfigure[]{
		\label{walkingP}
		\includegraphics[width=0.32\linewidth]{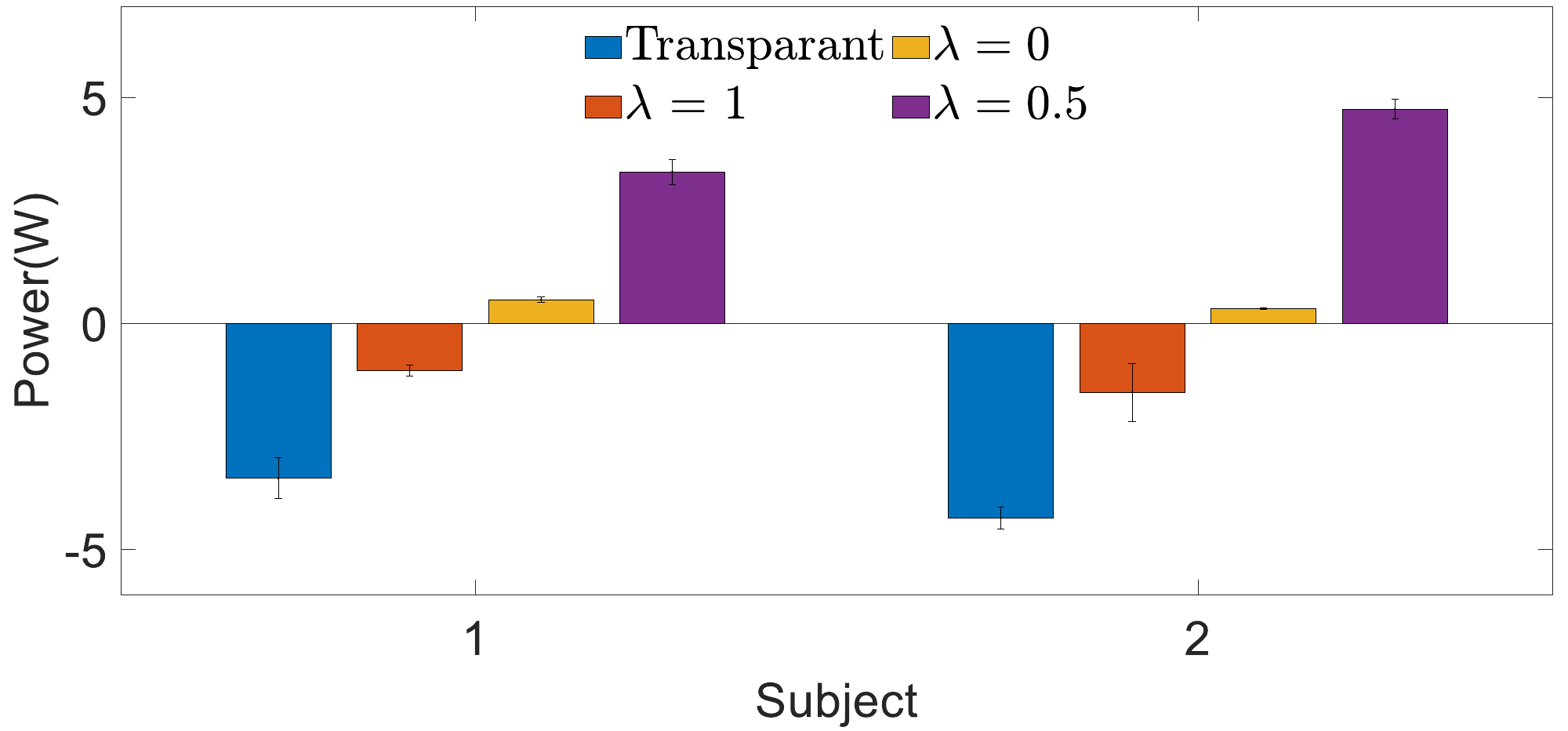}}
	\caption{Comparative evaluation of different modes across participants: (a) RMS of EMG signals from the tibialis anterior, indicative of user comfort; (b) RMS of EMG signals from the quadriceps femoris; (c) Average power transferred by the exoskeleton per step, representing the effectiveness of assistance.}
	\vspace{0cm}
	\label{exp_opt_walking}
\end{figure*}

To optimize the comfort of exoskeleton assistance, we recruited two participants not included in the pre-collected dataset for the application of the proposed HIL optimization. The attributes of these participants are summarized in Table \ref{participants_HIL}.
The participants were instructed to walk at a comfortable pace on a treadmill (FIT, Bertic, USA), and the frequency of reference trajectory was set to a constant according to (\ref{eq:omega}).
This speed remained constant during the optimization process, as depicted in Fig. \ref{snap_HIL}.

\begin{table}[h]
\caption{characteristics of subjects in HIL}
\centering
\begin{tabular}{ccccc} 
\toprule
Subject & Gender & Age(y) & Weight(kg) & Height(cm)  \\ 
\midrule
1 &  Male   & 23   & 65        & 172        \\
2   & Male   & 30   & 85        & 183        \\
\bottomrule
\end{tabular}
\label{participants_HIL}
\end{table}

The proposed variable impedance controller, described by (\ref{controlSection3}),(\ref{uf}) and (\ref{slow_controller}), was implemented in the exoskeleton robot to assist the human while detecting and relaxing physical conflicts. The impedance parameters were set as follows: $\bm C_d=15\bm I_2, \bm K_d=13\bm I_2$ where $\bm I_2$ is a $2\times 2$ identity matrix. The parameters of the weighting function were set as $\lambda_1=1, \chi_1=10, \chi_2=12$ and $\lambda_2=1$; and the control parameters were set as $\bm K_v=10^{-3}\bm I_2$ and $\bm K_z=25\bm I_2$. The experimental results are shown in Fig.\ref{Variable Impedance Weight}. The impedance parameters decreased (i.e., $w(s)$) when the anomaly score increased, relaxing the conflicts. 
The adjustment of impedance was continuous as the change in $w(s)$ was smooth, thereby ensuring safe human-robot interaction.
The impedance vector $\bm z$ remained zero, proving the realization of variable impedance.  
Leveraging the proposed anomaly-aware variable impedance controller, we aligned the trajectory improvement through Bayesian optimization in an online manner, without undermining the safety.

\begin{figure*}[!ht] 
	\centering 
	\vspace{-0.3cm} 
	\subfigtopskip=2pt 
	\subfigbottomskip=2pt 
	\subfigcapskip=-5pt 
	\subfigure[]{		\includegraphics[width=0.45\linewidth]{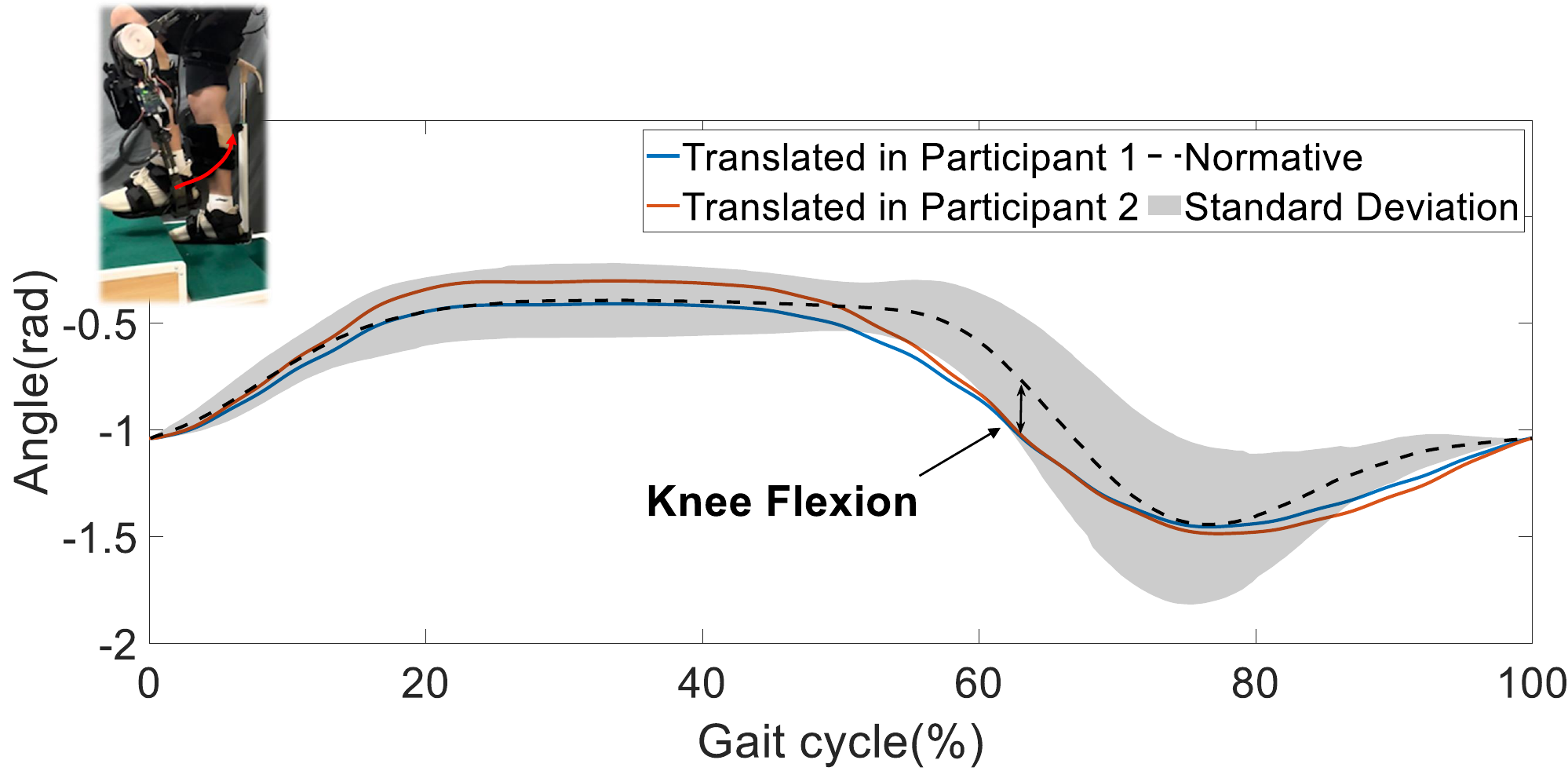}}
	\subfigure[]{
    \includegraphics[width=0.45\linewidth]{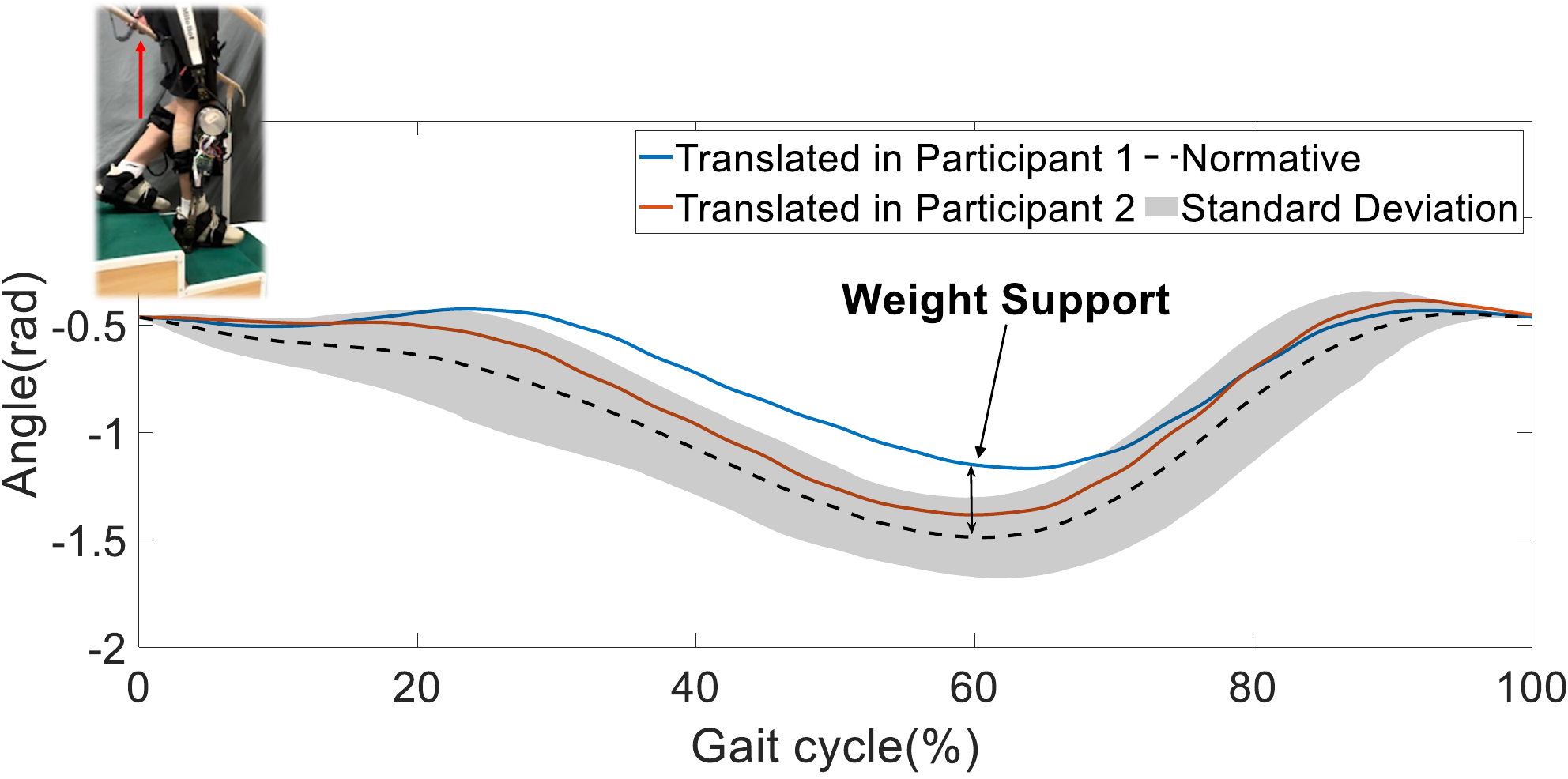}}\\
    \subfigure[]{	\includegraphics[width=0.45\linewidth]{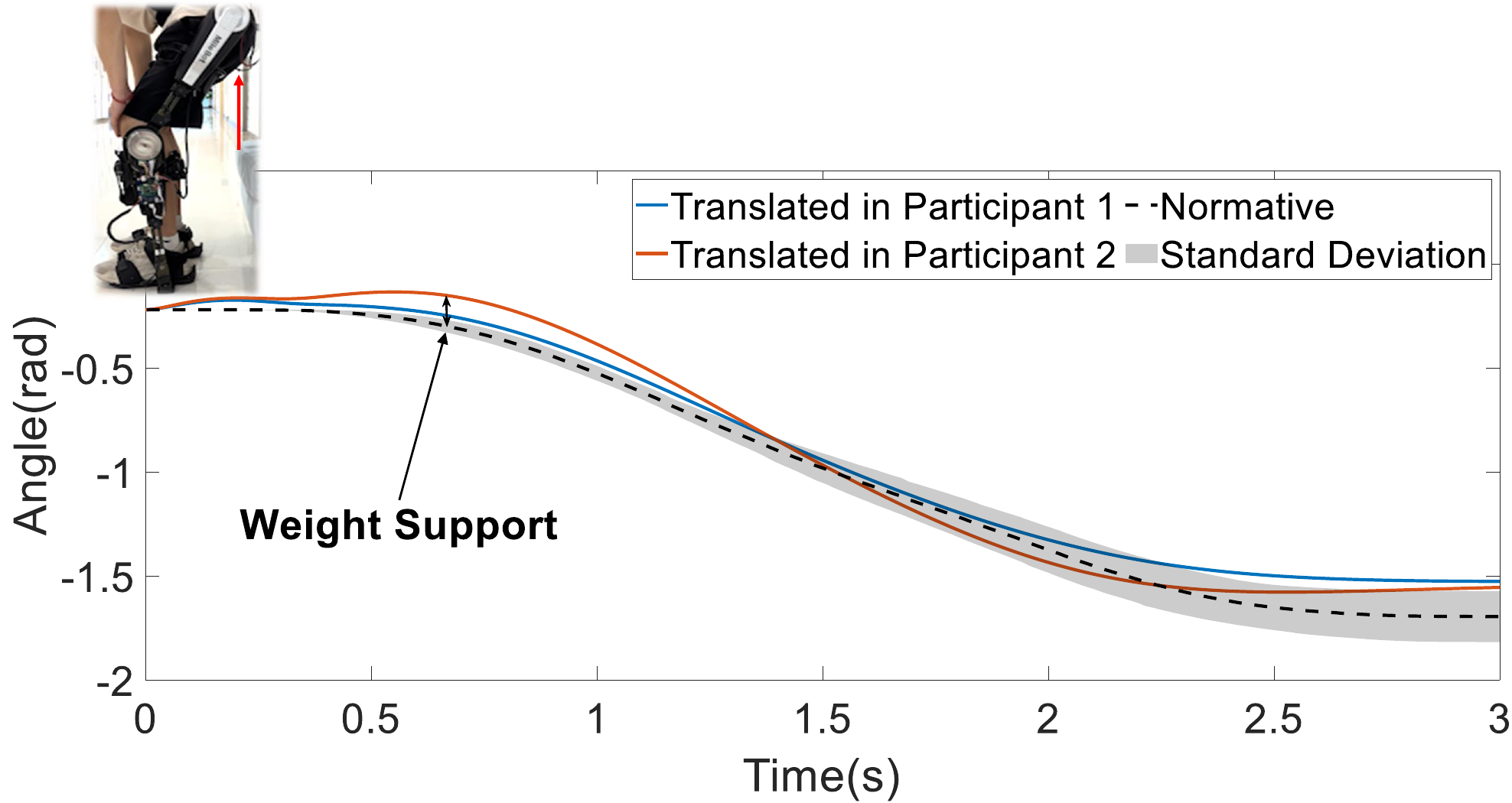}}
    \quad
    \subfigure[]{		\includegraphics[width=0.45\linewidth]{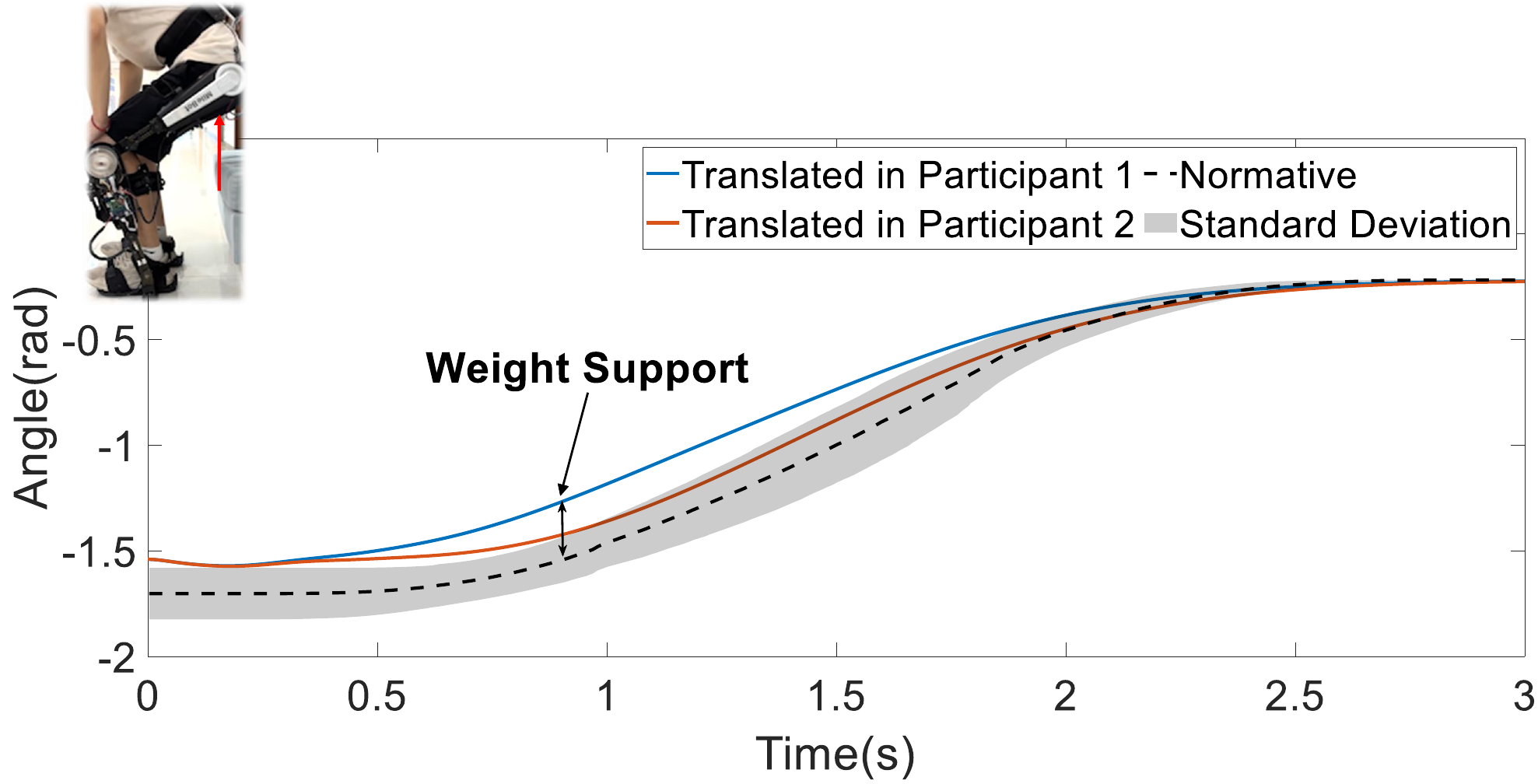}}
	\caption{Translation of optimized assistance across different tasks: (a) Ascending stairs; (b) Descending stairs; (c) Squatting; (d) Standing up.
 The red arrows denote the corresponding interactive forces exerted by the exoskeleton assistance.
 The blue and red solid lines represent the translation of optimized trajectories from Participants 1 and 2, respectively.
 The black dashed line represents the prediction from the task translator, with the surrounding black shade denoting its standard variation.}
	\vspace{0cm}
	\label{traj_tranferred}
\end{figure*}
To evaluate the performance of various participants in the walking task, we used the metric defined in (\ref{cost_function}) to guide trajectory optimization.
Only the motors in the knees were activated to aid the wearers, such that the weight matrix $\bm W$ was reduced to a simple weight vector.
During the initial phase of standing, the first $10$ elements of the weight vectors, which represent the encoded trajectory, were similar across all walking tasks. 
To alleviate the computational complexity, these elements were fixed to the mean of the data.
The assistive profile was updated upon every seventh heel strike. The initial six gaits were reserved for recording sensor signals to evaluate the performance with human interaction, and the final gait was input to the Bayesian optimization algorithm.
This approach prevented any sudden changes in the assistive profile that may have affected the balance of the wearers during consistent walking.

To accelerate optimization, we confined the parameter boundaries to $30\%$ around the identified optimal vector. 
Subsequently, we halved the boundaries whenever the optimal vector remained static over the last 10 steps. 
This method allowed the optimization process to refine the optimal weights, thereby enhancing the search efficiency without compromising the convergence certainty.
Optimization for each walking task required approximately 14 min, and 10 random restarts were implemented to bypass local minima.

Fig. \ref{HIL_graph} illustrates the HIL optimization outcomes. The blue and black solid lines correspond to the optimal trajectories when $\lambda = 1$ and $0$, respectively; the red trajectory depicts the mean and variance of the optimization outcomes for $\lambda$ values between $(0,1)$, i.e., $\lambda = 0.25,0.5,0.75$.
The results revealed that an optimal trajectory with $\lambda=1$ tended towards lazy assistance because the optimization term overly emphasized trajectory alignment, thereby causing the trajectory to converge to a curve with a smaller stretch amplitude.
In contrast, for $\lambda=0$, where the cost function consisted solely of the anomaly scores output by the NN, the final optimized trajectory exhibited heightened and frequent fluctuations.
When $\lambda$ ranged between 0 and 1, its effect on the trajectory optimization was significantly similar. 
This may be attributable to $\lambda$ eliminating the gap between the dataset and participant actions during optimizing based on the anomaly score.
Essentially, the first term in (\ref{cost_function}) was optimized to a lower value, and then the trajectory details were improved according to the human machine interaction results, thereby minimizing conflict and enhancing comfort.

Additional tests were performed to evaluate the effectiveness of the optimization results. 
Specifically, an indirect, quantitative evaluation of user comfort and assistance effectiveness was performed using EMG signals from the TA and QF, along with the average power transferred by the exoskeleton per step.
Fig. \ref{exp_opt_walking} shows the experimental outcomes in the transparent mode, with the optimized trajectory and without exoskeleton wear.
Considering the minimal influence of $\lambda$ on the trajectory optimization results, we included only cases where $\lambda$ equaled 0, 0.5, and 1.
The outcomes suggested that a trajectory produced with $\lambda = 0.5$ could not only ensure optimal comfort, as evidenced by the small amplitude of the EMG signal, but also provide effective assistance, i.e., maximum power transfer under all conditions.


\subsection{Translated Assistance}
\begin{figure*}[ht] 
	\centering 
	\vspace{-0.3cm} 
	\subfigtopskip=2pt 
	\subfigbottomskip=2pt 
	\subfigcapskip=-5pt 
	\subfigure[]{
		\label{walkingTA}
		\includegraphics[width=0.32\linewidth]{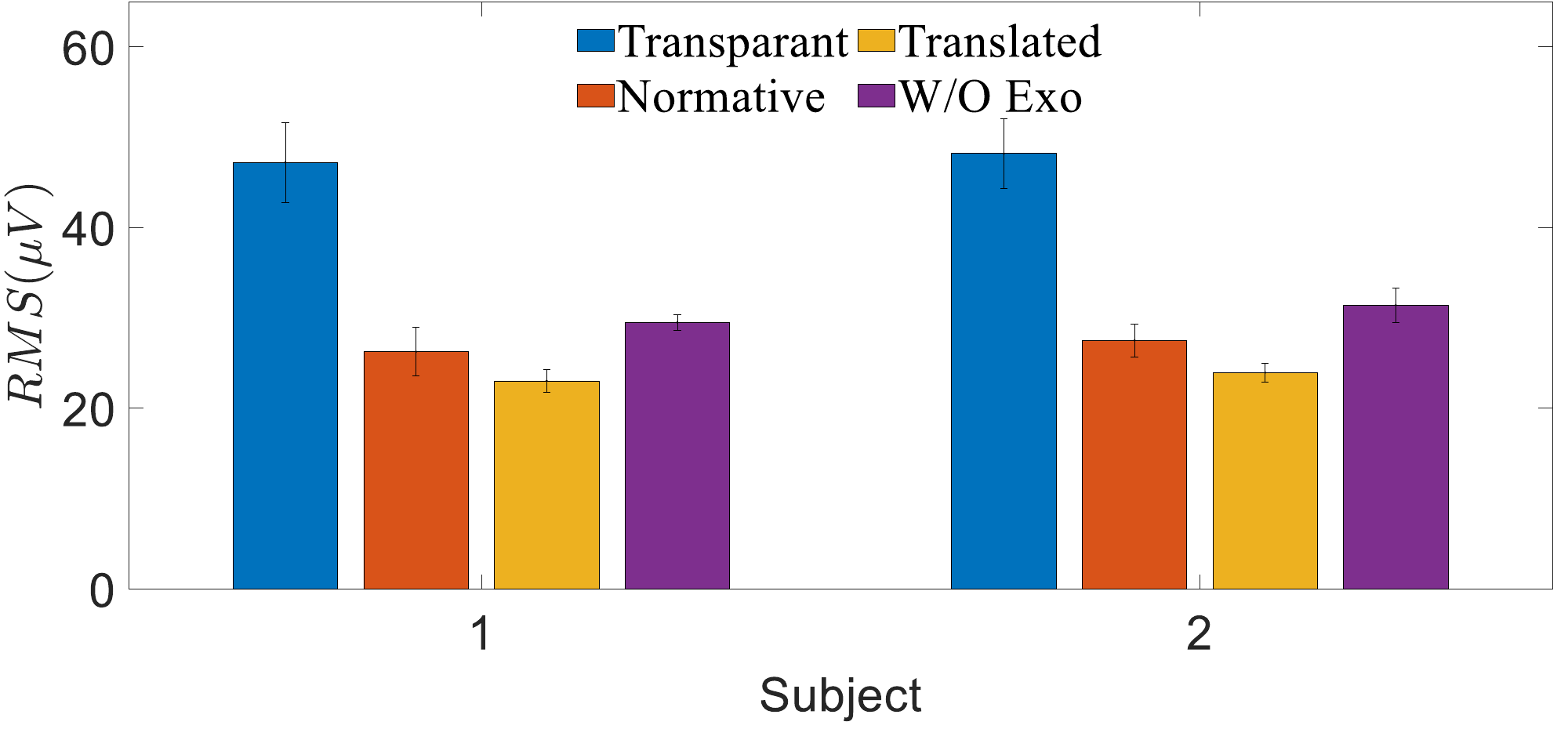}}
	\subfigure[]{
		\label{walkingQF}
		\includegraphics[width=0.32\linewidth]{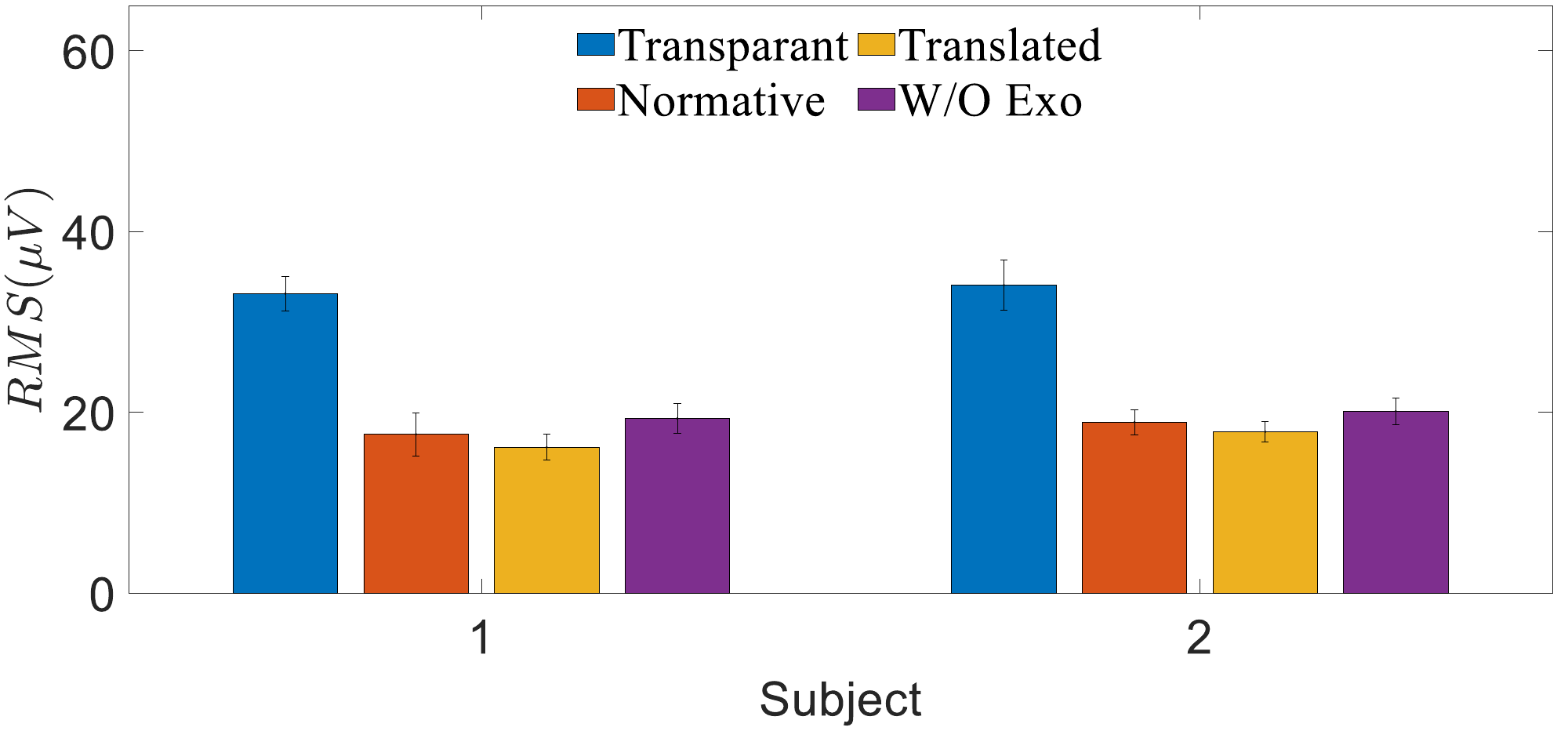}}
	\subfigure[]{
		\label{walkingP}
		\includegraphics[width=0.32\linewidth]{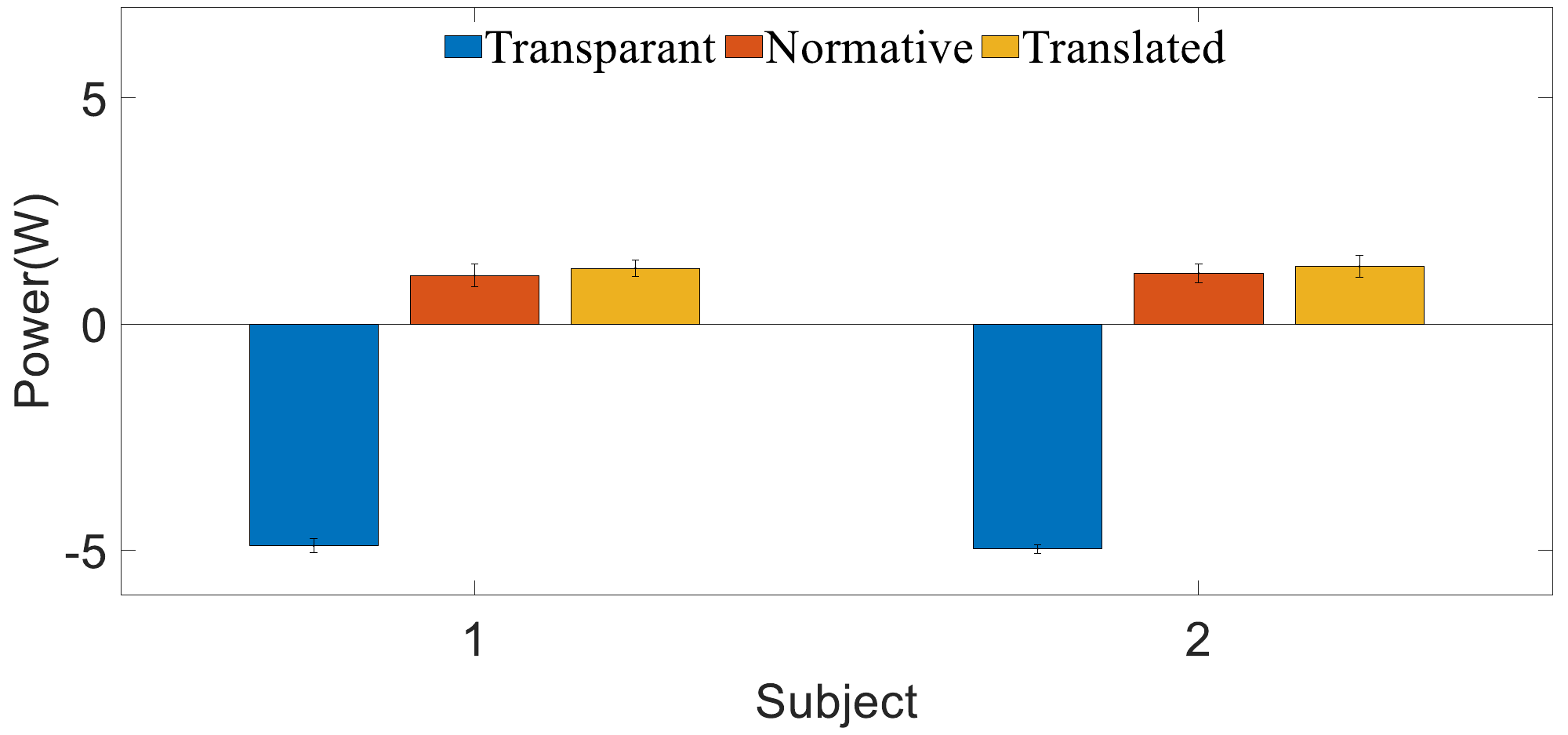}}
        \subfigure[]{
		\includegraphics[width=0.32\linewidth]{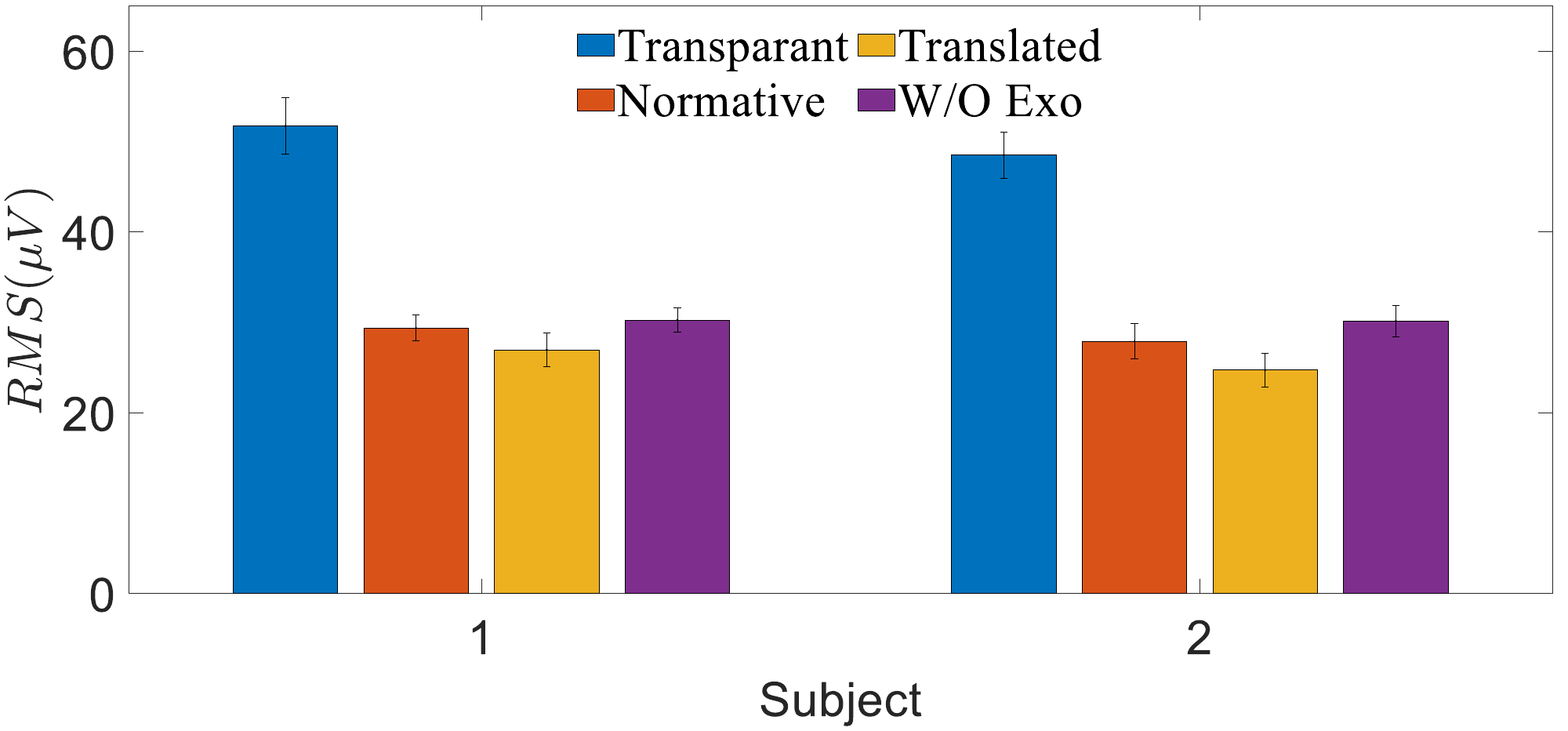}}
	\subfigure[]{
		\includegraphics[width=0.32\linewidth]{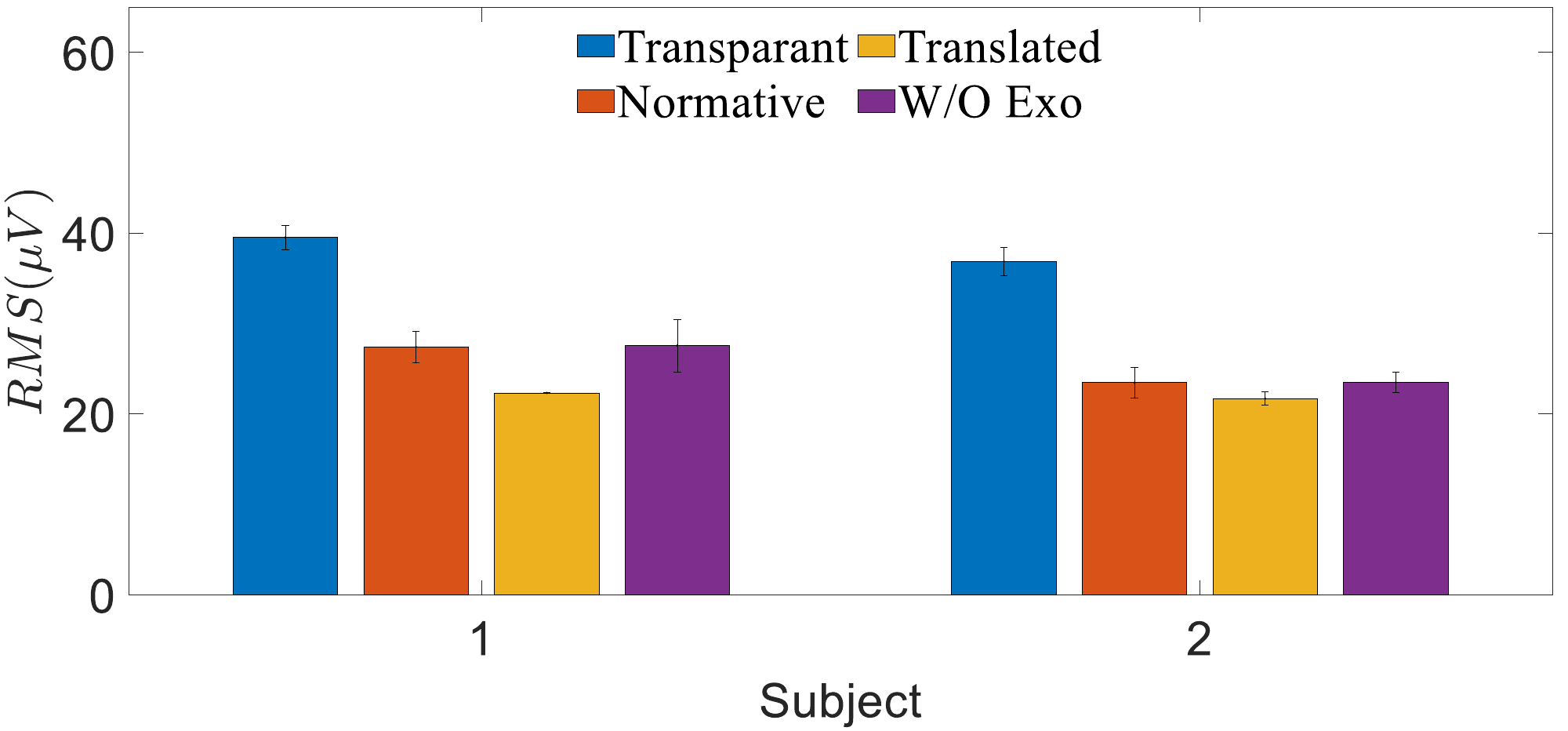}}
	\subfigure[]{
		\includegraphics[width=0.32\linewidth]{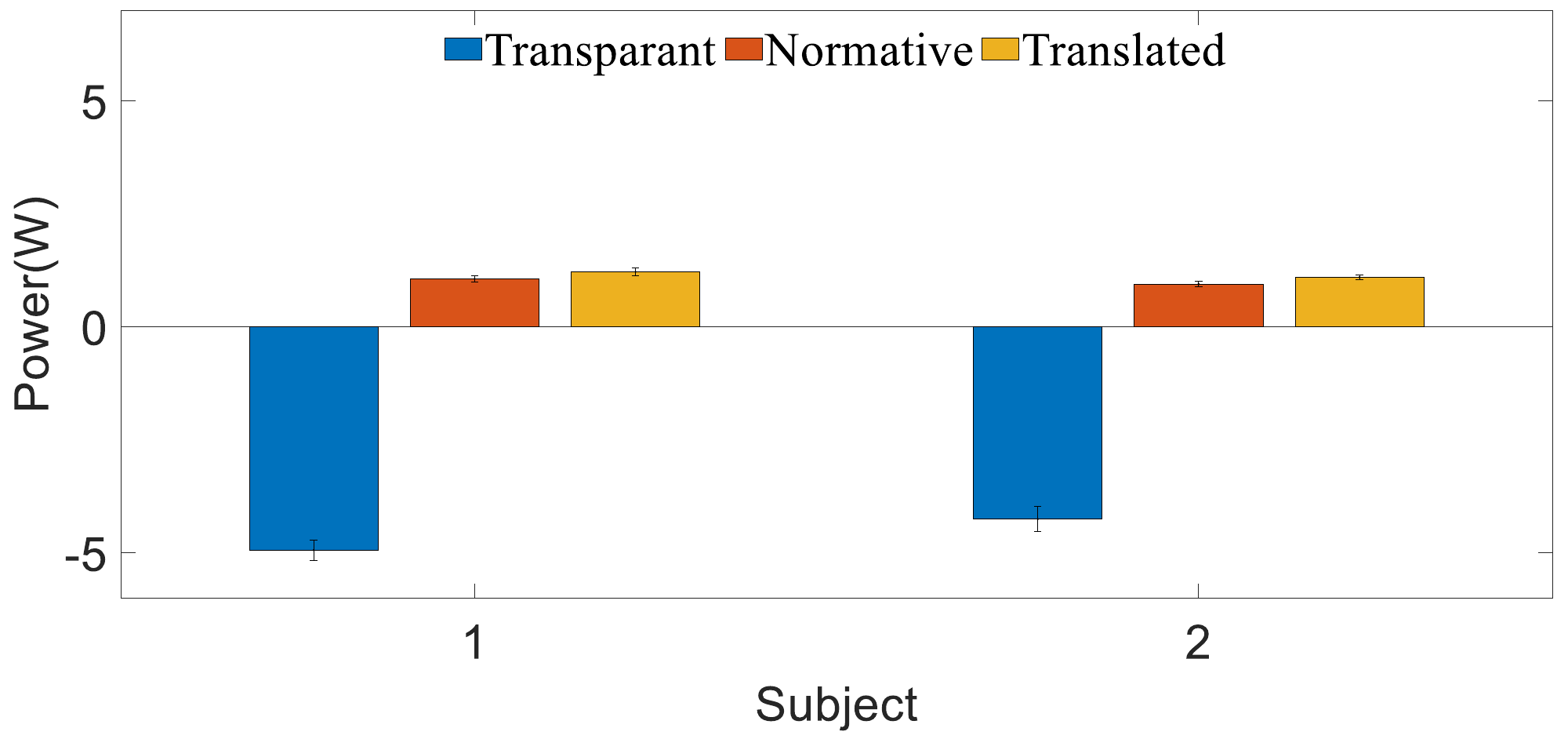}}
        \subfigure[]{
		\includegraphics[width=0.32\linewidth]{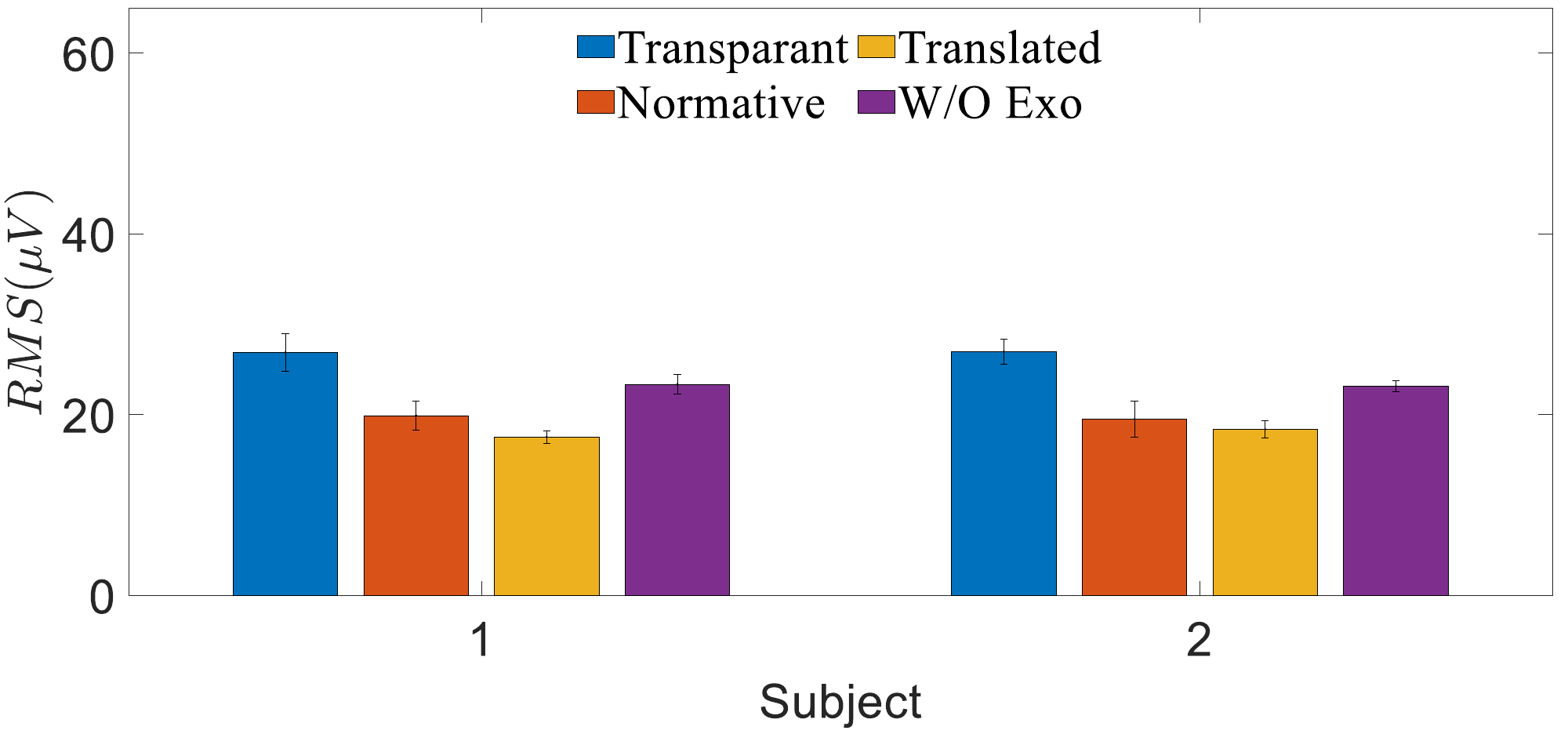}}
	\subfigure[]{
		\includegraphics[width=0.32\linewidth]{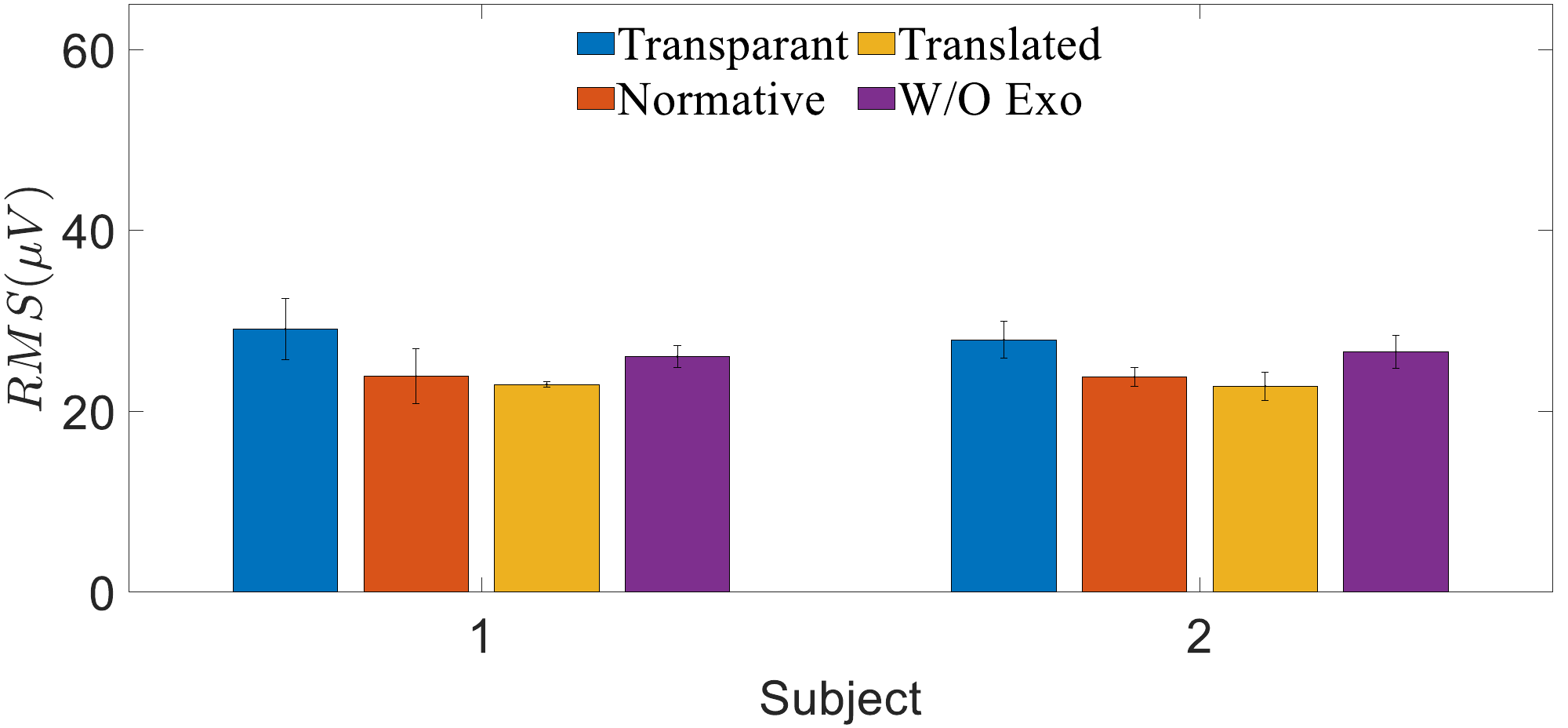}}
	\subfigure[]{
		\includegraphics[width=0.32\linewidth]{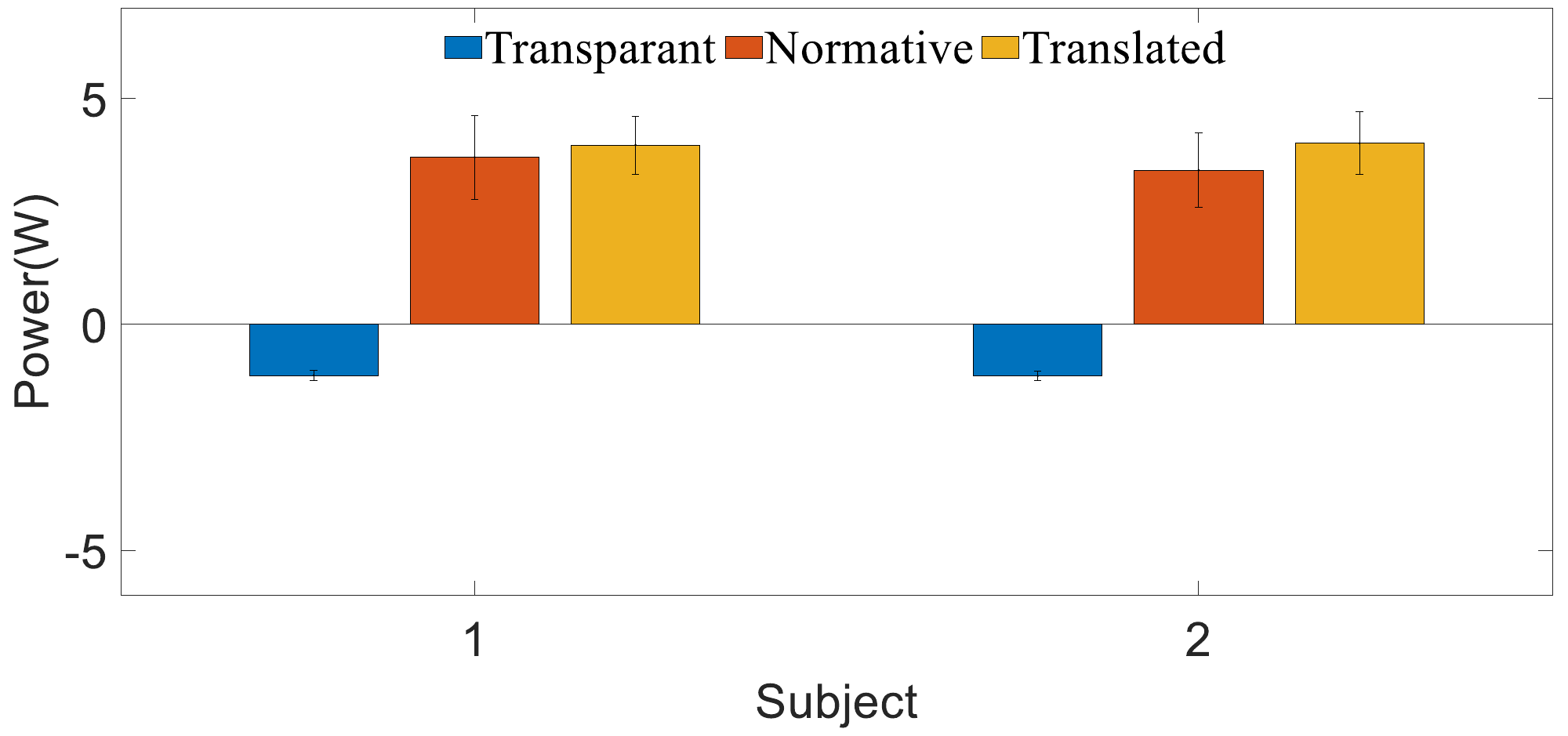}}
	\caption{Comparative evaluation of different modes across participants. Translation results for (a)-(c) stair ascent; 
 (d)-(f) stair descent;
 (g)-(i) squatting and standing up; Each column depicts different metrics:
 RMS of EMG signals from Tibialis Anterior and RMS of EMG signals from Quadriceps Femoris, indicating user comfort; and average power transferred by the exoskeleton per step, representing the effectiveness of assistance.}
	\vspace{0cm}
	\label{exp_opt_up}
\end{figure*}
The trajectories derived from HIL optimization were used to accommodate various tasks,
including ascending stairs, descending stairs, and the combined tasks of squatting and standing up.
Fig. \ref{traj_tranferred} offers a visualization of these translated trajectories along with the trajectory acquired in the transparent mode. 
Within our dataset, the average trajectory is referred to as the "normative trajectory".

The findings demonstrated that, compared with the normative trajectories, the translated trajectories could enable the provision of appropriate assistance. 
When ascending stairs, the elevated stepping leg received assistance in knee flexion to facilitate foot landing. 
Conversely, for descending stairs, the translated trajectories were consistent with the desired trajectory, which allows the robot to support the body weight through the impedance controller.
Similarly, tasks such as squatting and standing up were assisted by weight support achieved through trajectory optimization.
At the beginning of the squatting task, the translated trajectories exhibited an upward curve that deviated from the immediate decline observed in the normative trajectory, suggestive of preparatory actions for weight support.

To validate the effectiveness of the translated assistance, we compared the EMG signals of the TA and QF and the average power transfer per step under different conditions: transparent mode, normative mode, optimized mode, and a scenario without the exoskeleton, as summarized in Table \ref{Mode_desc}. 

\begin{table}[h]
\caption{Modes}
\centering
\begin{tabular}{l|l}
\hline
\textbf{Mode} & \textbf{Character} \\ \hline
Transparent &  Unrestricted mobility without apparent resistance. \\ \hline
Normative & $\bm q_d$ is the mean trajectory from the database. \\ \hline
Optimized & $\bm q_d$ is the translation result of the optimized trajectory. \\ \hline
W/O Exo & Unrestricted mobility without assistance. \\ \hline
\end{tabular}
\label{Mode_desc}
\vspace{-0.2cm}
\end{table}

The findings from these experiments are illustrated in Fig. \ref{exp_opt_up}.
Notably, the tasks of squatting and standing up were treated as one combined task, with a one-second separation used to distinguish the two.
In terms of minimizing the amplitude of the EMG signal and augmenting power transfer, the translated assistance demonstrated a modest improvement over the normative mode. 
This outcome suggested that the optimized and translated trajectories converged in the same latent space. 
Specifically, the translator, which was trained using the collected dataset, could be directly applied to the results of the HIL optimization.
Our proposed task translator could thus enhance comfort and assistance across various tasks based on the HIL results from the walking task. 
Therefore, we not only optimized the task trajectories that posed challenges for online optimization but also reduced the time required for task transfer.

\section{Discussion and Conclusions}

This paper proposes a novel adaptation framework for exoskeleton robots to customize the assistance provided to different wearers during multiple tasks. Specifically, the individualized trajectory is generated using DMP and online optimization, which can then be transformed to other trajectories for different tasks with the trained NN.
The individualized trajectory is incorporated in the variable impedance controller to drive the robot to provide assistance while ensuring a safe interaction. 
The advantages of the proposed framework can be summarized as follows: 1) an anomaly detection network is applied to improve wearing comfort during HIL optimization and relax potential conflicts during implementation; 2) the proposed approach is data-efficient with HIL adaptation ability, allowing the robot to promptly adapt to new users in new tasks, even with limited samples; and 3) the framework requires only proprioceptive sensors only and is thus easy to implement. 
The proposed method is evaluated using the training samples of 10 users and validation samples of two new users. Experiments and comparative studies are performed over four tasks (i.e., walking, stair ascent, stair descent, and sit-to-stand) to demonstrate the effectiveness of the exoskeleton robot. 
However, this study has several limitations.
Because optimization of the encoding trajectory involves multiple dimensions, we focused on optimizing assistance only for the knee joints to reduce time cost.
Moreover, the dataset is limited.
Thus, although the translation outcomes are enhanced, the magnitude of improvement remains modest. These limitations will be addressed in future work.

{\small
\bibliographystyle{ref/IEEEtran}
\bibliography{ref/IEEEabrv, ref/ref}
}

\end{document}